\pdfoutput=1

\documentclass[11pt]{article}
\usepackage[]{EMNLP2022}
\usepackage{times}
\usepackage{latexsym}
\usepackage{amsmath}
\usepackage[T1]{fontenc}
\usepackage[utf8]{inputenc}
\usepackage{microtype}
\usepackage{graphicx}
\usepackage{multirow}
\usepackage{CJK}
\usepackage{amsfonts,amssymb}
\usepackage{algorithm}
\usepackage{algpseudocode}
\usepackage{booktabs} 
\usepackage{subfigure}
\usepackage{bm}
\usepackage{marvosym}
\usepackage{enumitem}
\newcommand{\ctext}[1]{\begin{CJK}{UTF8}{gbsn}\small{#1}\end{CJK}}

\usepackage{xspace}
\newcommand{\sysname}{\texttt{RoChBert}\xspace}

\title{\sysname: Towards Robust BERT Fine-tuning for Chinese}

\author{
Zihan Zhang$^1$\thanks{\quad Equal contribution.}\qquad  
Jinfeng Li$^2$\footnotemark[1]\qquad 
Ning Shi$^{2,3}$\qquad 
Bo Yuan$^2$\qquad 
Xiangyu Liu$^2$\\
{\bf Rong Zhang$^2$\qquad 
Hui Xue$^2$\qquad 
Donghong Sun$^1$\qquad 
Chao Zhang$^1$\thanks{\quad  Corresponding author.}} \\
$^1$Tsinghua University; BNRist; Zhongguancun Lab\qquad $^2$Alibaba Group \\ $^3$Alberta Machine Intelligence Institute, Dept. of Computing Science, University of Alberta\\
\texttt{zhangzih19@tsinghua.org.cn, ning.shi@ualberta.ca}\\
\texttt{\{jinfengli.ljf, qiufu.yb, eason.lxy, stone.zhangr}\\
\texttt{hui.xueh\}@alibaba-inc.com,}\texttt{\{chaoz, sundonghong\}@tsinghua.edu.cn}
}

\begin{document}
\maketitle

\begin{abstract}

Despite of the superb performance on a wide range of tasks, pre-trained language models (e.g., BERT) have been proved vulnerable to adversarial texts.
In this paper, we present \sysname, a framework to build more Robust BERT-based models by utilizing a more comprehensive adversarial graph to fuse Chinese phonetic and glyph features into pre-trained representations during fine-tuning. Inspired by curriculum learning, we further propose to augment the training dataset with adversarial texts in combination with intermediate samples.
Extensive experiments demonstrate that \sysname outperforms previous methods in significant ways: 
(i) \textit{robust} -- \sysname greatly improves the model robustness without sacrificing accuracy on benign texts. Specifically, the defense lowers the success rates of unlimited and limited attacks by 59.43\% and 39.33\% respectively, while remaining accuracy of 93.30\%;
(ii) \textit{flexible} -- \sysname can easily extend to various language models to solve different downstream tasks with excellent performance; and 
(iii) \textit{efficient} -- \sysname can be directly applied to the fine-tuning stage without pre-training language model from scratch, and the proposed data augmentation method is also low-cost.\footnote{Our code will be available at \url{https://github.com/zzh-z/RoChBERT}.}

\end{abstract}
\section{Introduction}
The emergence of pre-trained language models (PLMs) has revolutionized natural language processing (NLP) to a new era. 
As a result, large-scale PLMs like BERT~\cite{devlin2018bert} have become the mainstream models for various downstream tasks including text classification~\cite{sun2019fine}, question answering~\cite{herzig2020tapas} and machine translation~\cite{zhu2020incorporating}, and have drastically boosted their performance.
In the Chinese domain, a wide variety of classical NLP tasks also benefit from the BERT-based PLMs.

\begin{figure}[t]
	\centering
    \includegraphics[width=.45\textwidth]{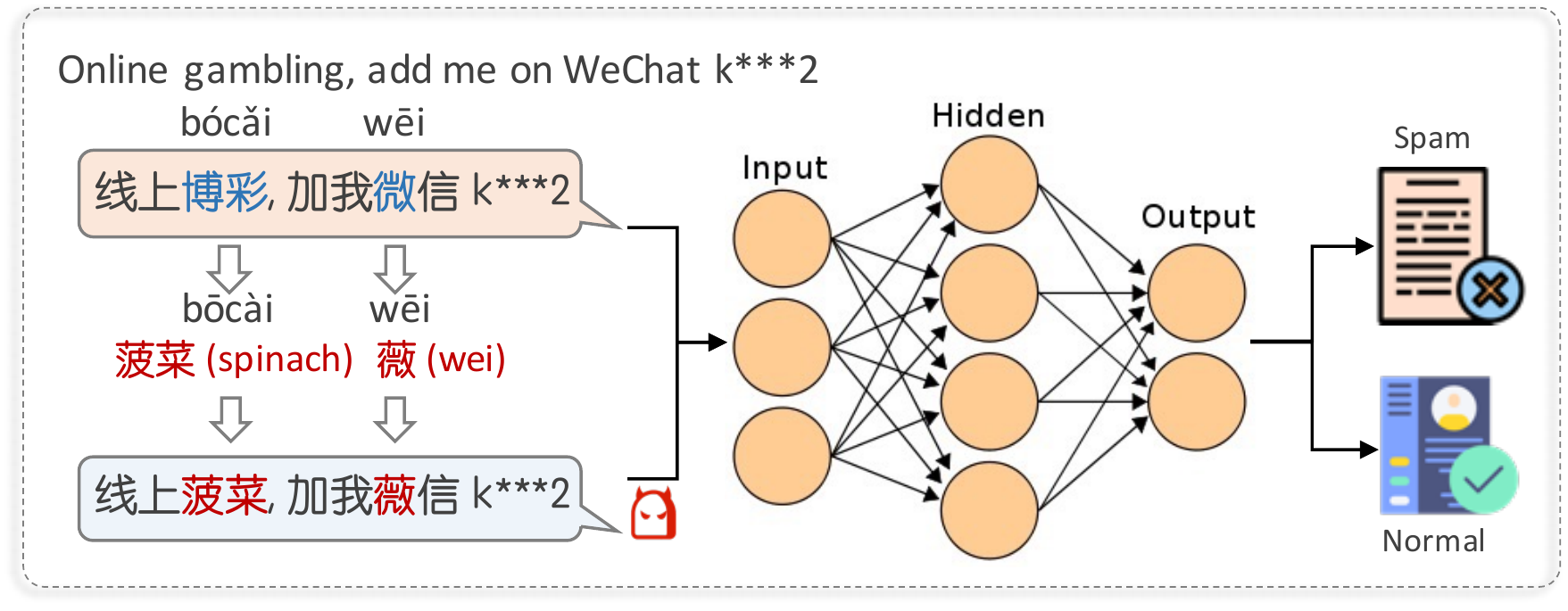}
    \caption{An example of Chinese adversarial text. ``\ctext{博彩}'' and ``\ctext{菠菜}'' share the same pronunciation of "bo cai".  ``\ctext{微}'' and ``\ctext{薇}'' share the same pronunciation of "wei" and similar components of ``\ctext{微}''.} 
    \label{fig:example}	
\end{figure}

Despite the impressive performance, BERT-based models have been proved to be vulnerable to maliciously generated adversarial texts~\cite{li2018textbugger,garg2020bae,li2020bert}.
Meanwhile, in the real-world scenarios, adversaries usually generate obfuscated texts, i.e., manually crafted adversarial texts to bypass online security-sensitive systems, which has posed severe physical threats to the deployed systems.
In contrast to the alphabet languages such as English, the meaning of individual Chinese character can be implied from its pronunciation and glyph.
Thus, substituting characters with others similar in characteristics can hardly bring impact to the context understanding.
Hence, unlike replacing words with synonyms to mislead English models, adversaries targeting Chinese models prefer substituting characters with others sharing similar pronunciation or glyph, as illustrated in Figure~\ref{fig:example}.

Several defense methods such as adversarial training (AT)~\cite{si2021better} and adversarial detection~\cite{bao2021defending} have been proposed recently. However, most of them focus on solving English adversarial texts. 
Based on Chinese characteristics, some Chinese-specific methods are proposed.
For example, ChineseBERT~\cite{sun2021chinesebert} incorporates Chinese pronunciation and glyph features into model pre-training, and it has achieved SOTA performance on many Chinese NLP tasks.
But ChineseBERT needs to be pre-trained from scratch and its number of parameters is quite large. It has to query the corresponding pinyin and glyph of input characters at each inference, which makes the training and prediction very slow.

To solve the challenges above, we propose \sysname, a lightweight and flexible method to strengthen robustness for Chinese BERT-based models.
Firstly, we generate an updated adversarial graph based on AdvGraph~\cite{li2021enhancing} to capture the phonetic and glyph relationships between characters, which are commonly exploited by adversarial texts.
As current AdvGraph only incorporates 3,000 commonly used characters and fails to capture certain glyph relationships, we update it to cover more characters and more glyph relationships.
Then, we use adversarial graph to learn the representation of characters based on Chinese pronunciation and glyph. The graph mainly serves as embedding weights within \sysname.
Meanwhile, we leverage the target model's hidden states of last layer as pre-trained representation.
These two kinds of embedding will then be fused in the fine-tuning of target model with specified downstream NLP tasks.
Secondly, we design a novel data augmentation method inspired by curriculum learning~\cite{bengio2009curriculum} for better fusion, which is proved to be more efficient than the traditional AT.
Specifically, we add both intermediate and adversarial texts into training datasets, which is very computationally efficient and will not decrease the accuracy on benign texts compared with traditional AT.
To the best of our knowledge, this is the first work to strengthen the robustness of pre-trained Chinese BERT-based models during fine-tuning.

Our contributions are summarized as follows:
\begin{enumerate}[leftmargin=*,noitemsep,topsep=0pt]
\item We present \sysname, a plug-in method to strengthen the robustness of BERT-based models during fine-tuning by incorporating adversarial knowledge. 
\item We update the adversarial graph and design an efficient data augmentation method, which considers intermediate samples for more effective use of adversarial texts.
\item Extensive experiments show that \sysname can drastically improve the robustness of target models against both known and unknown attacks without impacting the performance on benign texts. In addition, \sysname is flexible and efficient as it can be applied to most pre-trained models with specified downstream tasks while both fine-tuning and data augmentation are low-cost.
\end{enumerate}

\vspace{-0.2cm}
\section{Related Work}

\textbf{Pre-trained language model.} 
The applications of PLMs have achieved great success on various downstream NLP tasks and avoid training a new model from scratch.
BERT is first introduced to learn universal language representations via masked language model objective and next sentence prediction task, which is then improved in the following works such as RoBERTa~\cite{liu2019roberta} and ALBERT~\cite{lan2019albert}.
In the Chinese NLP domain, \citet{sun2021chinesebert} proposed ChinseBERT, which incorporates both the glyph and pinyin features of Chinese characters into language model pre-training, and also gained SOTA performance on many Chinese NLP tasks.
However, these works mainly focus on improving the model performance on benign texts, and the effort of enhancing model robustness is fairly limited.

\begin{figure*}[t]
	\centering
    \includegraphics[width=.95\textwidth]{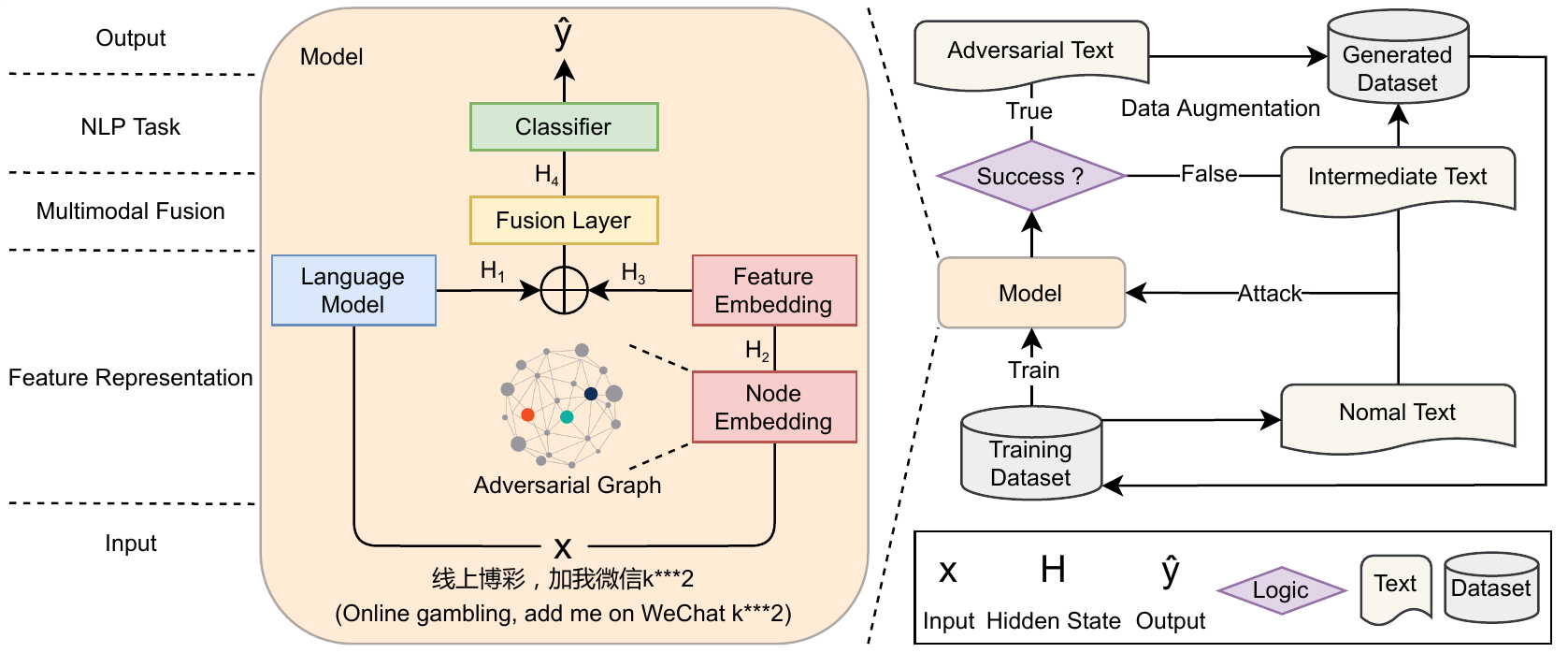}
    \caption{Overview of \sysname. The left part is the model architecture, and the right part is its workflow.}
    \label{fig:overview}		
\end{figure*}

\textbf{Exploration on the model robustness.} 
To further explore the vulnerabilities of NLP models in the real adversarial scenarios, a plenty of black-box attacks have been proposed under the practical assumption that adversary only has query access to the target models without any model knowledge~\cite{li2018textbugger,ren2019generating,garg2020bae}. 
To defend against such attacks, countermeasures like AT and adversarial detection have been proposed to mitigate the inherent model vulnerability.
Concretely, AT usually refers to retraining the target model by mixing adversarial texts into the original training dataset, which could be viewed as a kind of data augmentation~\cite{si2021better,ng2020ssmba}.
Adversarial detection is to check whether the input contains spelling errors or adversarial perturbation, and then restore it to the benign counterparts~\cite{zhang2020spelling,bao2021defending}.
Such methods both exhibit great efficacy in the English NLP domain, but they are hard to extend to the Chinese domain directly due to the language differences.
Hence, many studies have tried to design specific defense in terms of the unique property of Chinese.
For instance, \citet{wang2018hybrid} and \citet{cheng2020spellgcn} improved the Chinese-specific spelling check using the phonetic and glyph information.
\citet{li2021enhancing} proposed AdvGraph that involves an undirected graph to model the phonetic and glyph adversarial relationships among Chinese characters and improves the robustness of several traditional models.
\cite{su2022rocbert} proposed RoCBERT to enhance the robustness by pre-training the model from scratch with adversarial texts covering combinations of various Chinese-specific attacks, which may not be maintained in the downstream tasks.

\section{Methodology}

\vspace{-0.2cm}

Given a benign input text $x\in\mathcal{X}$ with the ground-truth label $y\in\mathcal{Y}$, a classifier $\mathcal{F}$ has learned the mapping $f\!:\mathcal{X}\!\rightarrow\!\mathcal{Y}$, which maps $x$ from the feature space $\mathcal{X}$ to the label space $\mathcal{Y}$ as $\mathcal{F}_{f}(x)=y$.
The adversary intend to obtain an adversarial text $x^{*}$ by adding some perturbations $\Delta x$ on $x$ with ${\Vert \Delta x\Vert}_{p}<\epsilon $, and deceive $\mathcal{F}$ into making a wrong prediction on $x^{*}$, i.e., $\mathcal{F}_{f}(x^{*})={\hat{y} }$ and $y\neq \hat{y} $. In this paper, we aim to improve the inherent robustness of $\mathcal{F}$ so to get a more robust model $\mathcal{F}'$ such that $\mathcal{F}'$ can resist the adversarial text $x^{*}$, i.e., $\mathcal{F}'_{f}(x^{*})=y$.

\textbf{Overview.}
As illustrated in Figure~\ref{fig:overview}, we first generate an updated adversarial graph based on phonetic and glyph relationships between all commonly used Chinese characters.
Then, we adopt node2vec~\cite{grover2016node2vec} to learn the representation of each character $x_{i}$,
which will then be used as the node embedding.
For $x_{i}$ in text $x$, its node embedding is concatenated and passed through feature extraction module.
Simultaneously, $x$ is also fed into the target PLM.
We utilize the hidden states of last layer as the pre-trained representation of $x$.
After concatenating pre-trained representation and feature embedding, we use a multimodal fusion module to further fuse the information from two channels.
Finally, the fused representation can be used for most of the downstream tasks.
Moreover, to better enhance the fusion process, we have designed an efficient data augmentation method, which makes full use of adversarial texts and helps the model further understand which characters may be similar in pronunciation and glyph.

\subsection{Adversarial Graph}
Adversarial graph is first presented in~\cite{li2021enhancing} as AdvGraph to improve robustness of classic deep learning model (e.g., TextCNN, LSTM). In AdvGraph, each node represents a Chinese character and each edge represents the phonetic or glyph relationship between two nodes.

However, the aforementioned AdvGraph only covers 3,000 commonly used characters. 
Moreover, the glyph relation in AdvGraph is constructed heavily relied on a g-CNN model which first converts each character into a image and then extracts the visual feature as glyph representation. 
It hence pays more attention to characters with similar structures, which will affect the representation learning of those characters with dissimilar structures. 
For instance, ``\ctext{微}'' and ``\ctext{徵}'' is considered to be closer by g-CNN since they are both left-middle-right radical structure, while ``\ctext{微}'' and ``\ctext{薇}'' is considered dissimilar since ``\ctext{薇}'' is up-down radical structure. 
This is because when converting to images, different radicals may cause the position of characters to shift, which will have a great impact on CNN.

To tackle the above problem, we intend to advance the adversarial graph as follows. 
First, we adopt the algorithm in Argot~\cite{zhang2020argot} to generate similar candidate sets for each character, with a combination of candidates in AdvGraph and StoneSkipping~\cite{jiang2019detect}. 
Next, we utilize stroke code to calculate similarity score for candidates.
Stroke refers to a line that is continuously written at a time, which is the smallest unit of Chinese characters.
Each number in stroke code represents a specific stroke, so it can be used to capture the same components between two characters.
Concretely, we calculate the \emph{longest common substring} between two codes as a part of our similarity score in addition to the distance of representation learned in g-CNN.
For each character in candidate set, once the similarity score exceeds the given threshold, we assume them as the similar pair.
Then, we add an edge between two characters for each similar pair, indicating they share similar pronunciation or glyph.

Finally, our adversarial graph incorporates 7,707 characters, and it has 109,706 edges, 32,894 of which is phonetic relationship, and the 81,108 is glyph relationship.
Note that the Chinese pronunciation, glyph and their relationships in our adversarial graph are general knowledge. 
We model the adversarial relationship in advance so that \sysname can protect target models against both known and unknown attacks.

\subsection{Multimodal Fusion}
In AdvGraph, graph embedding and semantic embedding are simply concatenated to generate fused representation and there is almost no interaction between these two kinds of feature.
Hence, we design a multimodal fusion module to make them more fully integrated during training process.

We first feed $x$ of length $l$ into PLM with hidden size $d_{1}$ to obtain pre-trained presentation.
Assume the target language model as the function $\mathcal{F}_{LM}$, we get the hidden states $H_{1}$ of its last layer:

\begin{equation}
 H_{1}=\mathcal{F}_{LM}(x)\in \mathbb{R}^{l\times d_{1}}.   
\end{equation}

Then we pass $x$ into Node Embedding Layer $\mathcal{F}_{G}$ whose weights are learned from adversarial graph to get $H_{2}$. We freeze the weights of $\mathcal{F}_{G}$ to preserve the independent phonetic and glyph features without being impacted by semantic information during training. The size of the vector $H_{2}$ is $d_{2}$, which is decided by the parameter of \emph{node2vec} algorithm:
\begin{equation}
H_{2}=\mathcal{F}_{G}(x)\in \mathbb{R}^{l\times d_{2}}.
\end{equation}
Since the focus of model should be different on the benign and adversarial text, we introduce Transformer Encoder layer $\mathcal{T}_{1}$ to help model capture the differences.
There still exists the \emph{out-of-vocabulary} problem in the node embedding though we have updated the original AdvGraph, since if the character is not similar with any character or its frequency is too low, it will not be included in the graph.  
In addition, node embedding doesn't incorporate English letters, numbers, punctuation, and special symbols, it will be hard to align with pre-trained embedding $H_{1}$ and it may even bring negative effects. 
We use flattening to preserve its phonetic and glyph features and alleviate the problem that some features may be too sparse caused by \emph{out-of-vocabulary}. Then we pass it into a linear layer with weights $\bm{W}$ and bias $b$ to map it into a vector of size $d_{1}$ as sequence feature representation $H_{3}$:
\begin{equation}
H_{3}=\bm{W}^\top (\mathcal{T}_{1}(H_{2}))+\bm{b}\in \mathbb{R}^{d_{1}}.
\end{equation} 
In order to facilitate the subsequent concatenation, we repeat the vector $l$ times to obtain $H'_{3}\in \mathbb{R}^{l\times d_{1}}$  as the final feature embedding.

We first fuse these two kinds of embedding by concatenating $H_{1}$ and $H'_{3}$.
To make them better interact with each other during training, we apply Transformer Encoder layer $\mathcal{T}_{2}$ for the fusion. The final embedding $H_{4}$ of input $x$ is given by:
\begin{equation}
H_{4}=\mathcal{T}_{2}(H_{1}\oplus H'_{3})\in \mathbb{R}^{l\times d_{1}}.
\end{equation}
Finally, we flatten $H_{4}$  and feed it into a classifier to get output $\hat{y}\in\mathcal{Y} $.

\begin{algorithm}[t]
    \small
      \caption{: The detail of data augmentation.}
      \label{alg:dataaug}
      \begin{algorithmic}[1]
        \Require
        Training dataset $D$, the target classifier $\mathcal{F}$ with mapping $f$.
        \Ensure
          New training dataset $D_{ag}$.
        \For{$x \in D$}
        \State $tmp\gets \{\}$
            \State $\hat{y}=\mathcal{F}_{f}(x)$
            \If{$\hat{y}$ is not the ground-truth label}
                \State continue
            \EndIf
            \State $x^{*}=x$, $\hat{y}^{*}=\hat{y}$
            \While{$\hat{y}^{*}==\hat{y}$}
                \State $x^{*}=x^{*}+\Delta x$ \Comment{According to attack algorithms}
                \State $\hat{y}^{*}=\mathcal{F}_{f}(x^{*})$
                \State $tmp\gets tmp \bigcup \{x^{*}\}$
                \If{all the words in $x$ are modified}
                    \State break
                \EndIf
            \EndWhile
            \If{$\hat{y}^{*}\neq \hat{y}$ and ${\Vert x^{*}-x \Vert}_{p}<\epsilon_{max} $}
                \State $D_{ag}\gets D_{ag}\bigcup tmp$
            \EndIf
            \If{$size(D_{ag})>size(D)$}
                \State $D_{ag}\gets D_{ag}\bigcup D$
                \State break
            \EndIf
        \EndFor
        \label{code:glyph:candidates_end}
        \State \Return $D_{ag}$;
      \end{algorithmic}
    \end{algorithm}

\subsection{Efficient Data Augmentation}
\label{sec:dataaug}

It has been reported that data augmentation can improve the robustness of models~\cite{si2021better,ng2020ssmba}. 
The intuitive method is to retrain the target model by mixing the generated adversarial texts into the original training dataset.
However, most of the conventional AT methods are usually very cost-expensive.
For instance, in the black-box attacks, it often requires an extremely large amount of queries for accessing the target models to determine the attack direction in each iteration when generating just one adversarial text.

Conventional AT typically involves the entire dataset to conduct attacks while only collecting successful adversarial samples for augmentation.
To efficiently utilize the generated texts and obtain more samples in the shortest time, we have made some improvements to the traditional AT. 

For generating adversarial texts with fewest perturbations,
there are various strategies designed in the attacks to determine the added perturbation in each iteration.
Usually, this perturbation can cause the most significant drop on the output probability of its original label, which is most hopeful to mislead the target model within the next iterations among all the candidate generated texts.
Therefore, during the attack, many intermediate texts will be generated before the final adversarial text is generated.
Although the labels of these intermediate texts are still correct, their confidences have declined to varying degrees.
Hence, to some extend, in addition to adversarial texts, the intermediate texts can also help models make firm decisions.

We generate augmented data based on the training dataset $D$.
For each $x$ in $D$, we perturb it and collect the intermediate texts $tmp$. If the generated $x^{*}$ can deceive the model $\mathcal{F}_{f}$, we will incorporate $tmp$ (the final generated $x^{*}$ is also in $tmp$) into new dataset $D_{ag}$.
Specifically, to ensure the quality of texts, we will limit the modification rate $\epsilon_{max}$ during the generation process.
If the size of $D_{ag}$ is larger than the size of $D$, we assume we have got enough texts and early stop the process.
Finally, we add $D$ into $D_{ag}$ and $D_{ag}$ will be our new training dataset.
The details of the algorithm are presented in Algorithm~\ref{alg:dataaug}, in which $\epsilon_{max}$ and $size(D_{ag})$ have the following relationship:
\begin{equation}
\nonumber 
\begin{split}
size(D_{ag})&\leq l_{x_{1}^{*}}\times \epsilon_1+l_{x_{2}^{*}}\times \epsilon_2+...+l_{x_{n}^{*}}\times \epsilon_n,\\
&\mathrm{ s.t. }\quad \epsilon_i\leq\epsilon_{max},i\in 1...n
\end{split}
\end{equation}

$l_{x_{i}^{*}}$ denotes the length of adversarial text $x_{i}^{*}$. $n$ denotes the number of generated adversarial texts. As the base of perturbation is word, we can get:
\begin{equation}
    \begin{split}
        size(D_{ag})&\leq (l_{x_{1}^{*}}+l_{x_{2}^{*}}+...+l_{x_{n}^{*}})\times \epsilon_{max}\\
        &\leq l_{avg}\times n\times\epsilon_{max}.
    \end{split}
\end{equation}
Therefore we have:
\begin{equation}
n\geq\frac{size(D_{ag})}{l_{avg}\times\epsilon_{max}}.
\end{equation}
So $n$ increases as $l_{avg}$ and $\epsilon_{max}$ decrease.
In the augmentation stage, more used adversarial texts ($n$ is bigger) will result in better model robustness,
but the more computation is needed.
If we set small $\epsilon_{max}$, we will get high quality adversarial texts. 
They will promote the model robustness with less impact on accuracy when added into training dataset.
On the contrary, when we set big $\epsilon_{max}$, the adversarial texts we need to generate become fewer, which reflects the high efficiency of \sysname.

So when the accuracy of the target model is relatively high and the average length of texts is long, we recommend to use smaller  $\epsilon_{max}$ to make $n$ bigger and improve robustness. If the average length of texts is short, $n$ will increase, so we urge to use bigger $\epsilon_{max}$ for efficiency boosting.
And when the model accuracy is not very high, we suggest to use smaller $\epsilon_{max}$ to keep the accuracy as high as possible.

Our method inherits the main idea of curriculum training~\cite{bengio2009curriculum}, and the model can learn the features of the adversarial text at different stages. Moreover, it can improve model's robustness without compromising the accuracy on benign texts, which can be proved in Section~\ref{sec:clean_perf}.

\section{Experiment}

\begin{table}[t]
    \centering
    \small
    \resizebox{\linewidth}{!}{%
    \begin{tabular}{lcccc}
    \toprule
    \textbf{Model}          & \textbf{Chnsenti.} & \textbf{DMSC} & \textbf{THUC.} & \textbf{OCNLI} \\
    \cmidrule(r){1-1}
    \cmidrule(r){2-5}
    ChineseBERT             & 95.25                & 92.95    & 97.87  &\textbf{73.20}   \\
    \cmidrule(r){1-1}
    \cmidrule(r){2-5}
    BERT\textsubscript{base}       & 95.33                & 93.02   & 98.07  &71.57    \\
    +SC       & 94.42                & 92.85   & 98.07&70.57      \\
    +\sysname (PWWS)           & 95.58                & 93.05    & 97.87 &  67.76   \\
    +\sysname (TextBugger)     & 95.83                & 92.75   & 98.00 & 67.34    \\
    +\sysname (Random)         & 95.92                & 92.70    & \textbf{98.13 } &70.25   \\
    \cmidrule(r){1-1}
    \cmidrule(r){2-5}
    BERT\textsubscript{wwm}        & 94.58                & 92.51   & 97.87 &70.33         \\
    +SC       & 93.58                & 92.45   & 97.93&69.08      \\
    +\sysname (PWWS)           & 94.92                    & 92.70   & 97.93 &67.23        \\
    +\sysname (TextBugger)     & 95.75                    & 93.30  & 97.80  &67.71         \\
    +\sysname (Random)         & 95.25                    & 91.45   & 98.00 &70.09         \\
    \cmidrule(r){1-1}
    \cmidrule(r){2-5}
    BERT\textsubscript{wwm/ext}    & \textbf{96.00}                & 93.29  & 97.73 &71.16          \\
    +SC       & 95.00                & 93.20   & 97.80    &70.68  \\
    +\sysname (PWWS)           & 95.58                    & \textbf{94.00}   & 97.80          &68.31\\
    +\sysname (TextBugger)     & 95.42                    & 93.30   & 97.87  &  68.65      \\
    +\sysname (Random)         & 95.83                    & 93.60   & 97.73  &71.40        \\
    \cmidrule(r){1-1}
    \cmidrule(r){2-5}
    RoBERTa\textsubscript{wwm/ext} & 95.58                & 92.89   & 98.00  &71.29        \\
    +SC        & 94.50                & 92.95   & 97.87   &70.88   \\
    +\sysname (PWWS)           & 95.58                    & 93.10  & \textbf{98.13 }  &68.31        \\
    +\sysname (TextBugger)     & 95.83                    & 93.56   & 97.93  &69.09        \\
    +\sysname (Random)         & 95.50                    & 93.45   & 98.07   &72.45      \\ 
    \bottomrule
    \end{tabular}
    }
    \caption{Model performances on benign texts.}
    \label{tab:clean-perf}
    \end{table}

\begin{table*}[!t]
    \renewcommand{\arraystretch}{1.05}
    \centering
    \small
    \resizebox{\linewidth}{!}{%
    
    \begin{tabular}{cccccccccc}
    \toprule
    \multicolumn{1}{l}{\multirow{2}{*}{\textbf{Model}}} & \multicolumn{3}{c}{\textbf{PWWS}}                                  & \multicolumn{3}{c}{\textbf{TextBugger}}                            & \multicolumn{3}{c}{\textbf{Random}}            \\
    \cmidrule(r){2-4}
    \cmidrule(r){5-7}
    \cmidrule(r){8-10}
    \multicolumn{1}{c}{}                                & \textbf{UASR} & \textbf{LASR} & \multicolumn{1}{c}{\textbf{MR}} & \textbf{UASR} & \textbf{LASR} & \multicolumn{1}{c}{\textbf{MR}} & \textbf{UASR} & \textbf{LASR} & \textbf{MR} \\
    \cmidrule(r){1-1}
    \cmidrule(r){2-10}
     & \multicolumn{9}{c}{{\tt ChnSetiCorp}} \\
    \cmidrule(r){1-1}
    \cmidrule(r){2-10}
    \multicolumn{1}{l}{ChineseBERT}                     & 79.73         & 40.97         & \multicolumn{1}{c}{27.22}          & 93.25         & 42.67         & \multicolumn{1}{c}{\textbf{23.64}}          & 54.91         & \textbf{3.38}          & \textbf{51.23}          \\
    \cmidrule(r){1-1}
    \cmidrule(r){2-4}
    \cmidrule(r){5-7}
    \cmidrule(r){8-10}
    \multicolumn{1}{l}{BERT\textsubscript{base}}               & 83.62         & 67.66         & \multicolumn{1}{c}{12.96}          & 97.45         & 69.26         & \multicolumn{1}{c}{16.12}          & 52.77         & 8.19          & 42.85          \\
    \multicolumn{1}{l}{+SC}               & 82.75         & 64.86         & \multicolumn{1}{c}{13.74}          & 96.49         & 71.25         & \multicolumn{1}{c}{14.85}          & 54.42         & 7.56          & 43.01          \\
    \multicolumn{1}{l}{+\sysname}                          & 65.18         & 31.63         & \multicolumn{1}{c}{29.49}          & 64.45         & 34.92         & \multicolumn{1}{c}{20.35}          & 39.49        & 10.98          & 37.48          \\
    \cmidrule(r){1-1}
    \cmidrule(r){2-4}
    \cmidrule(r){5-7}
    \cmidrule(r){8-10}
    \multicolumn{1}{l}{BERT\textsubscript{wwm}}               & 87.53         & 56.56         & \multicolumn{1}{c}{19.86}          & 98.28         & 64.73         & \multicolumn{1}{c}{16.89}          & 48.27         & 6.45          & 44.91          \\
    \multicolumn{1}{l}{+SC}               & 84.62         & 57.42         & \multicolumn{1}{c}{18.75}          & 97.63         & 68.60         & \multicolumn{1}{c}{15.95}          & 50.54         & 5.59          & 45.62          \\
    \multicolumn{1}{l}{+\sysname}                          & 64.42         & 22.94        & \multicolumn{1}{c}{\textbf{41.94}}       & 62.24         & 35.23         & \multicolumn{1}{c}{19.6}     & 42.31     & 6.36             & 44.88              \\
    \cmidrule(r){1-1}
    \cmidrule(r){2-4}
    \cmidrule(r){5-7}
    \cmidrule(r){8-10}
    \multicolumn{1}{l}{BERT\textsubscript{wwm/ext}}               & 72.04         & 42.93         & \multicolumn{1}{c}{22.21}          & 92.93         &  53.27        & \multicolumn{1}{c}{20.16}          & 56.96         & 5.80          & 40.44          \\
    \multicolumn{1}{l}{+SC}               & 75.00         & 50.53         & \multicolumn{1}{c}{18.19}          & 90.93         & 57.59         & \multicolumn{1}{c}{17.47}          & 57.07         & 5.91          & 41.04          \\
    \multicolumn{1}{l}{+\sysname}                          & \textbf{62.75}       & 24.44             & \multicolumn{1}{c}{37.08}              & 65.74         & 31.70         & \multicolumn{1}{c}{23.17}  & \textbf{38.45}       & 6.45             & 44.26            \\
    \cmidrule(r){1-1}
    \cmidrule(r){2-4}
    \cmidrule(r){5-7}
    \cmidrule(r){8-10}
    \multicolumn{1}{l}{RoBERTa\textsubscript{wwm/ext}}               & 76.46         & 44.11         & \multicolumn{1}{c}{23.25}          & 99.78         & 54.19         & \multicolumn{1}{c}{21.49}          & 57.58         & 5.83          & 44.14          \\
    \multicolumn{1}{l}{+SC}               & 80.47         & 52.23         & \multicolumn{1}{c}{19.62}          & 98.20         & 59.66         & \multicolumn{1}{c}{18.26}          & 55.94         & 6.90          & 44.00          \\
    \multicolumn{1}{l}{+\sysname}                          & 65.85        & \textbf{22.69}             & \multicolumn{1}{c}{39.14}     & \textbf{54.18}          & \textbf{28.57}         & \multicolumn{1}{c}{20.49}          & 38.59       & 8.41           & 36.60              \\
    \cmidrule(r){1-1}
    \cmidrule(r){2-10}
     & \multicolumn{9}{c}{{\tt DMSC}} \\
    \cmidrule(r){1-1}
    \cmidrule(r){2-10}
    \multicolumn{1}{l}{ChineseBERT}                     & 78.76         & 60.35         & \multicolumn{1}{c}{16.64}          & 92.20         & 59.37         & \multicolumn{1}{c}{18.47}          & 53.30         & 7.04          & 48.75          \\
    \cmidrule(r){1-1}
    \cmidrule(r){2-4}
    \cmidrule(r){5-7}
    \cmidrule(r){8-10}
     \multicolumn{1}{l}{BERT\textsubscript{base}}               & 76.70         & 61.06         & \multicolumn{1}{c}{15.74}          & 78.75         & 60.19         & \multicolumn{1}{c}{13.79}          & 56.31         & 7.66          & 46.69          \\
    \multicolumn{1}{l}{+SC}               & 83.24         & 63.24         & \multicolumn{1}{c}{17.87}          & 82.49         & 63.24         & \multicolumn{1}{c}{13.54}          & 57.51         & 6.38          & 47.25          \\
    \multicolumn{1}{l}{+\sysname}                          & 68.67         & 46.70         & \multicolumn{1}{c}{23.66}          & \textbf{36.36}         & \textbf{29.22}         & \multicolumn{1}{c}{12.43}          & 44.94         & 10.13         & 39.21          \\
    \cmidrule(r){1-1}
    \cmidrule(r){2-4}
    \cmidrule(r){5-7}
    \cmidrule(r){8-10}
     \multicolumn{1}{l}{BERT\textsubscript{wwm}}               & 95.66         & 76.66         & \multicolumn{1}{c}{13.53}          & 99.67         & 76.33         & \multicolumn{1}{c}{13.92}          & 52.88         & 6.30          & 43.41          \\
    \multicolumn{1}{l}{+SC}               & 94.46      & 74.38         & \multicolumn{1}{c}{13.66}          & 98.91         & 77.96         & \multicolumn{1}{c}{13.07}          & 55.70         & 7.38          & 43.59          \\
    \multicolumn{1}{l}{+\sysname}                          & 64.94             & 50.00             & \multicolumn{1}{c}{14.00}      & 44.85      & 33.47            & \multicolumn{1}{c}{15.94}              & 46.78             & 11.91            & 40.22              \\
    \cmidrule(r){1-1}
    \cmidrule(r){2-4}
    \cmidrule(r){5-7}
    \cmidrule(r){8-10}
    \multicolumn{1}{l}{BERT\textsubscript{wwm/ext}}               & 85.52         & 60.30         & \multicolumn{1}{c}{17.63}          & 99.79         & 69.31         & \multicolumn{1}{c}{16.35}          & 57.51         & \textbf{5.90}          & \textbf{49.93}          \\
    \multicolumn{1}{l}{+SC}               & 88.41         & 63.95         & \multicolumn{1}{c}{16.98}          & 99.36         & 72.85         & \multicolumn{1}{c}{15.36}          & 59.12         & 6.76          & 48.85          \\
    \multicolumn{1}{l}{+\sysname}                          & 75.40             & 46.31             & \multicolumn{1}{c}{22.37}              & 40.36             & 29.98             & \multicolumn{1}{c}{14.65}              & \textbf{37.38}             & 6.87             & 47.99              \\
    \cmidrule(r){1-1}
    \cmidrule(r){2-4}
    \cmidrule(r){5-7}
    \cmidrule(r){8-10}
    \multicolumn{1}{l}{RoBERTa\textsubscript{wwm/ext}}               & 69.70         & 50.76         & \multicolumn{1}{c}{20.04}          & 83.12         & 53.90         & \multicolumn{1}{c}{18.69}          & 52.71        & 7.79          & 39.31          \\
    \multicolumn{1}{l}{+SC}               & 75.54         & 55.19         & \multicolumn{1}{c}{18.54}          & 85.82         & 57.47         & \multicolumn{1}{c}{17.57}          & 54.87         & 7.68          & 40.70          \\
    \multicolumn{1}{l}{+\sysname}                   & \textbf{55.15}     & \textbf{28.06}            & \multicolumn{1}{c}{\textbf{34.80}}              & 59.25             & 35.27            & \multicolumn{1}{c}{\textbf{19.52}}              & 44.86             & 7.92             & 43.98              \\ \bottomrule
    \end{tabular}
    
    \begin{tabular}{ccccccccc}
    \toprule
    \multicolumn{3}{c}{\textbf{PWWS}}                                  & \multicolumn{3}{c}{\textbf{TextBugger}}                            & \multicolumn{3}{c}{\textbf{Random}}            \\
    \cmidrule(r){1-3}
    \cmidrule(r){4-6}
    \cmidrule(r){7-9}
    \textbf{UASR} & \textbf{LASR} & \multicolumn{1}{c}{\textbf{MR}} & \textbf{UASR} & \textbf{LASR} & \multicolumn{1}{c}{\textbf{MR}} & \textbf{UASR} & \textbf{LASR} & \textbf{MR} \\
    \cmidrule(r){1-9}
    \multicolumn{9}{c}{{\tt THUCNews}} \\
    \cmidrule(r){1-9}
    71.55        & 23.44         & \multicolumn{1}{c}{44.82}          & 69.80         & 36.23         & \multicolumn{1}{c}{23.63}          & 78.40         & 1.13          & 64.68          \\
    \cmidrule(r){1-3}
    \cmidrule(r){4-6}
    \cmidrule(r){7-9}
    81.31         & 25.64         & \multicolumn{1}{c}{44.60}          & 58.43         & 43.21         & \multicolumn{1}{c}{14.87}          & 80.59         & 2.56          & 61.42          \\
    80.51         & 29.18         & \multicolumn{1}{c}{40.87}          & 59.90         & 46.02         & \multicolumn{1}{c}{13.26}          & 79.08         & 2.55          & 61.96          \\
    66.35        & 5.11         & \multicolumn{1}{c}{63.71}          & 9.66        & 9.10        & \multicolumn{1}{c}{8.05}          & 51.17       & 0.81        & \textbf{66.57}          \\
    \cmidrule(r){1-3}
    \cmidrule(r){4-6}
    \cmidrule(r){7-9}
    73.77	    & 36.27		    & \multicolumn{1}{c}{35.17}              & 78.28	        & 46.93           & \multicolumn{1}{c}{21.07}         & 74.90	       & 2.67	        & 58.81             \\
    76.66         & 30.19         & \multicolumn{1}{c}{38.37}          & 72.26         & 45.45         & \multicolumn{1}{c}{17.96}          & 73.49         & 2.97          & 58.68          \\
    76.60             & 5.19             & \multicolumn{1}{c}{\textbf{64.00}}      & 13.09      & 11.15            & \multicolumn{1}{c}{11.94}              & \textbf{45.77}             & 0.72            & 64.67              \\
    \cmidrule(r){1-3}
    \cmidrule(r){4-6}
    \cmidrule(r){7-9}
    79.63	        & 21.29             & \multicolumn{1}{c}{48.05}              & 86.39	        & 29.99	        & \multicolumn{1}{c}{\textbf{31.85}}              & 78.81	            &1.74	        &60.79             \\
    82.72         & 21.98         & \multicolumn{1}{c}{46.59}          & 79.04         & 33.54         & \multicolumn{1}{c}{26.08}          & 79.24         & 2.76          & 60.04          \\
    66.80            & 7.89             & \multicolumn{1}{c}{60.09}              & 13.10             & 8.80             & \multicolumn{1}{c}{16.31}              & 51.07            & \textbf{0.61}             & 66.04             \\
    \cmidrule(r){1-3}
    \cmidrule(r){4-6}
    \cmidrule(r){7-9}
    72.58        & 17.02	          & \multicolumn{1}{c}{50.12}              & 81.24	        & 30.27	             & \multicolumn{1}{c}{28.46}              & 79.51	        & 1.22	& 62.27           \\
    81.51         & 19.41         & \multicolumn{1}{c}{48.52}          & 77.22         & 35.55         & \multicolumn{1}{c}{24.33}          & 79.57         & 1.23          & 62.43          \\
    \textbf{59.92}     & \textbf{4.68}          & \multicolumn{1}{c}{63.54}              & \textbf{5.49}           & \textbf{4.88}            & \multicolumn{1}{c}{7.70}              &  59.04            & 1.83            & 60.33             \\
    \cmidrule(r){1-9}
    \multicolumn{9}{c}{{\tt OCNLI}} \\
    \cmidrule(r){1-9}
    62.57        & 46.22         & \multicolumn{1}{c}{17.32}          & 73.78         & 35.27         & \multicolumn{1}{c}{25.16}          & 38.92         & 8.38          & 40.92          \\
    \cmidrule(r){1-3}
    \cmidrule(r){4-6}
    \cmidrule(r){7-9}
    58.68         & 42.29         & \multicolumn{1}{c}{17.39}          & 65.84         & 35.95         & \multicolumn{1}{c}{22.96}          & 40.08         & 10.06          & 38.59          \\
    56.50         & 42.38         & \multicolumn{1}{c}{15.81}          & 65.73         & 36.08         & \multicolumn{1}{c}{21.87}          & 38.88         & 8.81          & 38.82          \\
    \textbf{43.58}        & \textbf{29.73}         & \multicolumn{1}{c}{17.90}          & \textbf{48.99}        & 24.21        & \multicolumn{1}{c}{24.53}          & \textbf{36.50}       & \textbf{5.65}        & \textbf{47.04}          \\
    \cmidrule(r){1-3}
    \cmidrule(r){4-6}
    \cmidrule(r){7-9}
    56.06	    & 40.42		    & \multicolumn{1}{c}{17.09}              & 64.23	        & 32.54           & \multicolumn{1}{c}{23.97}         & 36.34	       & 8.03	        & 42.91             \\
    55.33         & 40.68         & \multicolumn{1}{c}{16.28}          & 63.30         & 33.43         & \multicolumn{1}{c}{22.69}          & 36.56         & 8.68          & 41.33          \\
    50.00             & 30.18             & \multicolumn{1}{c}{20.79}      & 50.36      & 20.32            & \multicolumn{1}{c}{27.54}              & 39.21             & 7.05            & 46.48              \\
    \cmidrule(r){1-3}
    \cmidrule(r){4-6}
    \cmidrule(r){7-9}
    62.22	        & 45.56             & \multicolumn{1}{c}{16.46}              & 69.31	        & 37.08	        & \multicolumn{1}{c}{23.15}              & 40.69	            & 8.61	        & 41.58             \\
    61.79         & 47.70         & \multicolumn{1}{c}{15.26}          & 69.32         & 37.10         & \multicolumn{1}{c}{21.83}          & 43.10         & 10.74          & 40.65          \\
    51.75            & 32.75             & \multicolumn{1}{c}{21.27}              & 54.69             & \textbf{20.23}             & \multicolumn{1}{c}{\textbf{28.24}}              & 39.52            & 7.45             & 45.33             \\
    \cmidrule(r){1-3}
    \cmidrule(r){4-6}
    \cmidrule(r){7-9}
    66.21        & 46.02	          & \multicolumn{1}{c}{17.65}              & 80.08	        & 37.09	             & \multicolumn{1}{c}{26.35}              & 40.66	        & 7.01	& 43.34           \\
    65.10         & 47.31         & \multicolumn{1}{c}{17.48}          & 79.03         & 40.14         & \multicolumn{1}{c}{24.01}          & 41.66         & 8.83          & 40.91          \\
    60.14     & 35.11          & \multicolumn{1}{c}{\textbf{22.33}}              & 51.43           & 21.57            & \multicolumn{1}{c}{27.31}              &  36.86            & 7.70            & 42.60             \\ 
    \bottomrule
    \end{tabular}
    }
    \caption{Model performance against different attacks.}
    \label{tab:robustness}
    \end{table*}

\subsection{Experiment Settings}

\textbf{Datasets.}
We use three different tasks, \emph{sentiment analysis}, \emph{text classification} and \emph{natural language inference} to evaluate \sysname.
The datasets are ChnSentiCorp\footnote{\url{https://github.com/pengming617/bert_classification/tree/master/data}}, DMSC\footnote{\url{https://www.kaggle.com/utmhikari/doubanmovieshortcomments/}}, THUCNews~\cite{sun2016thulac} and OCNLI~\footnote{\url{https://github.com/cluebenchmark/OCNLI}}.
The details are shown in Appendix~\ref{sec:ap_dataset}.
For {\tt OCNLI}, we attack the two sentences, \emph{premise} and \emph{hypothesis} separately. 
Here, we only present the corresponding results of \emph{premise}, the other part is shown in Appendix~\ref{sec:ap_hypothesis}.

\textbf{Setup.} 
As the target models are Chinese BERT-based models and the perturbation strategies are quite different from those for English, we cannot use the specialized BERT attack methods like BERT-ATTACK~\cite{li2020bert}.
Hence, we utilize three widely used attacks, i.e., PWWS~\cite{ren2019generating}, TextBugger~\cite{li2018textbugger} and random attack in the black-box setting to evaluate the robustness of \sysname.
The details are shown in Appendix~\ref{sec:ap_attack}.
We leverage the characters with similar glyph or pronunciation to form words and substitute the original ones after locating the important parts by corresponding attack algorithms.
The attacks are conducted on 1,000 texts sampled from test set.
We use base BERT (BERT\textsubscript{base}), BERT trained with whole word masking strategy (BERT\textsubscript{wwm}) and with extended data (BERT\textsubscript{wwm/ext}), and  RoBERTa (RoBERTa\textsubscript{wwm/ext}) as the target models.\footnote{\url{https://huggingface.co}}
We also take ChineseBERT which incorporates pronunciation and glyph features into  pre-training, as a baseline.
In addition, we compare \sysname with the Chinese spelling corrector (SC)~\footnote{\url{https://github.com/shibing624/pycorrector}} method. 
In SC-based defenses, each input text is firstly restored by a corrector to eliminate errors in the text before being sent into the target model. For fair comparison, the adversarial texts against the SC baseline are generated by treating corrector and the model as a whole pipeline.
And $\epsilon_{max}$ in data augmentation are shown in Appendix~\ref{sec:ap_epsilon}.

\textbf{Metrics.} 
We utilize four metrics, i.e., accuracy, modification rate (MR), unlimited attack success rate (UASR) and limited attack success rate (LASR), to comprehensively evaluate \sysname.
Accuracy reflects model's generalization on benign texts.
MR reflects the average percentage of perturbed characters in adversarial texts under unlimited attack setting.
UASR means the percentage of texts that can generate adversarial texts successfully without any limitations.
To guarantee the quality of adversarial texts, the maximum MR will be 0.2 when calculating LASR.

\subsection{Model Performance}
\label{sec:clean_perf}
We first evaluate the model performances on benign texts since defense shouldn't compromise the natural generalization of models.
As summarized in Table~\ref{tab:clean-perf}, SC reduces the accuracy of models due to its own errors. 
It is observed that \sysname achieves relatively high accuracy, which is comparable with the original BERT-based models and ChineseBERT. In some cases, models defended by \sysname even outperform the baselines. This is mainly because the additional phonetic and glyph features can help model capture semantic information in some degree, which has been proved in ChineseBERT.
In addition, the gains on accuracy also reflect the key impact of data augmentation.

\subsection{Robustness Against Attack}
\textbf{Effectiveness.}
Then, we evaluate the efficacy and robustness of \sysname.
The results are shown in Table~\ref{tab:robustness}.
It is observed that the defend effect of SC on the target model is not obvious, which means that when it is combined with the target model as a whole, the attacker regards them as a black-box model to conduct attack, and can still easily generate adversarial texts.
ChineseBERT indeed enhance the robustness compared with original models, but the improvement is very limited while \sysname brings noticeable reduction on both UASR and LASR.
In the best case, UASR is decreased by 75.75\%  when \sysname is fine-tuned on RoBERTa\textsubscript{wwm/ext} with {\tt THUCNews} and attacked by TextBugger, and LASR is decreased by 42.86\% when it is fine-tuned on BERT\textsubscript{wwm} with {\tt DMSC} and attacked by TextBugger.
This indicates that \sysname can significantly weaken the attack and enhance robustness of models without sacrificing their performance.
Simultaneously, MR of adversarial texts generated against \sysname is higher, which means that attackers need to modify more words to deceive \sysname, and the generated texts will be harder to comprehend. In some cases, MR is decreased, this is because the UASR has dropped significantly, adversaries fail to attack even if they modify the texts extensively.

Then, we analyze the relation between MR and ASR by setting the maximum MR from 0.1 to 0.5.
The results of TextBugger attack on {\tt THUCNews} are shown in Figure~\ref{fig:thu_textbugger}. The other results are in Appendix~\ref{sec:ap1}.
The dashed, dotted and solid lines in the same color represent the target model, the model protected by SC and \sysname respectively.
It proves that \sysname retains robustness against attack facing with the change of MR.
Compared with baselines, ASR of models protected by \sysname increases slightly as the MR grows,
indicating that \sysname can significantly enhance model robustness even in adaptive settings.

The above experiments have shown the robustness of \sysname against known attacks as it will use the specified attack algorithm during data augmentation.
Here we conduct another experiment to demonstrate \sysname 's performance against unknown attack.
We leverage TextBugger to generate adversarial texts and their immediate texts since we have known that TextBugger is the strongest attack among the three attacks as shown in Table~\ref{tab:robustness}.
And we assume that using the strongest attack during data augmentation can strengthen the robustness of the model against unknown attacks as much as possible.
As shown in Table~\ref{tab:unknown-attack}, we can see that both UASR and LASR have decreased greatly while MR increases, indicating that when facing unknown attacks, \sysname still has the ability to defend. 
Hence we can add defense to the target model in advance without knowing the specific adversaries. 
Obviously, we can leverage other stronger or multiple attacks during data augmentation.
\begin{figure}[t]
	\centering
    \includegraphics[width=.45\textwidth]{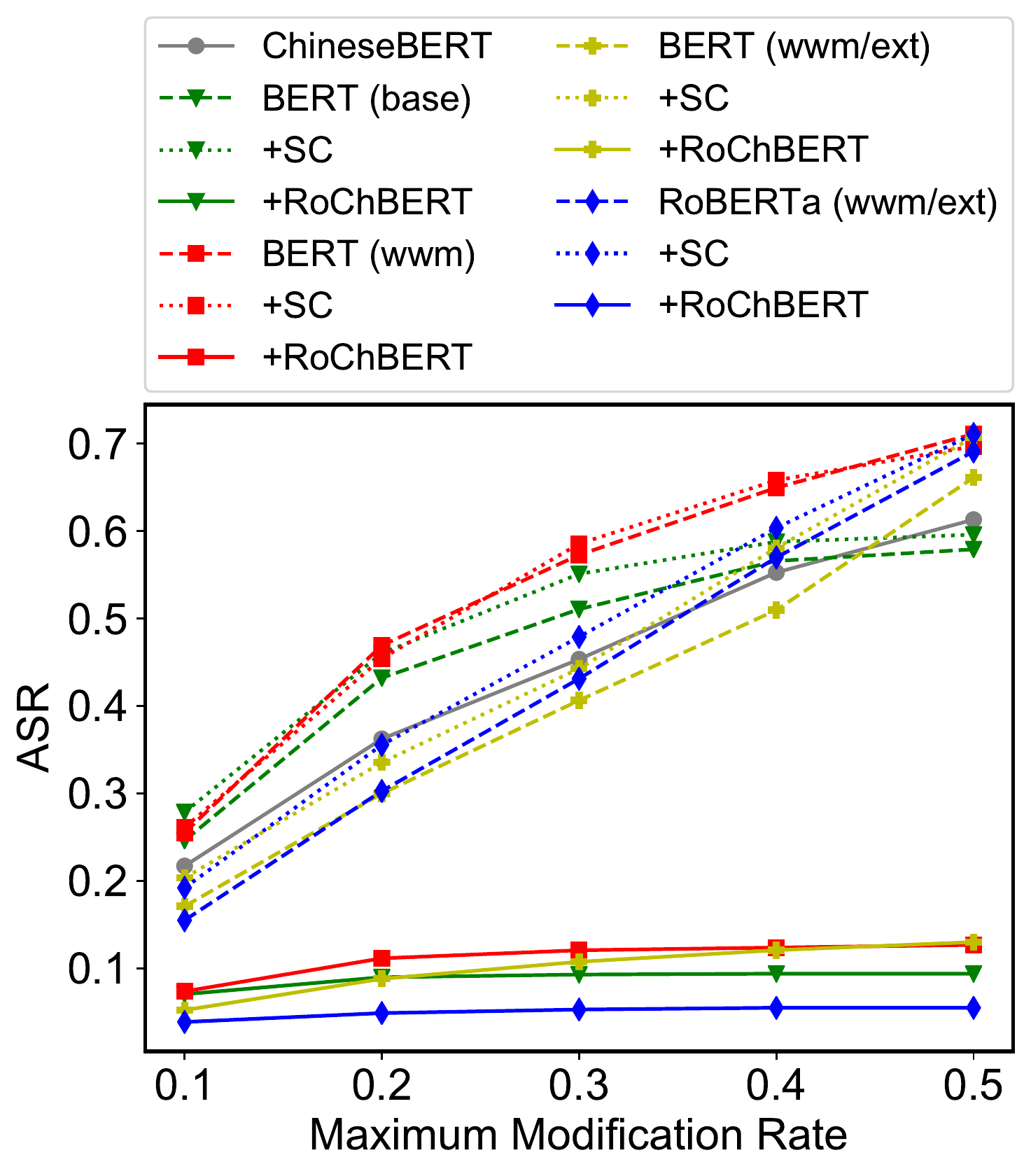}
    \caption{Relation between maximum MR and ASR. }
    \label{fig:thu_textbugger}		
\end{figure}
\begin{table}[t]
    \renewcommand{\arraystretch}{1.05}
    \centering
    \small
    \resizebox{0.85\linewidth}{!}{%
    \begin{tabular}{llccc}
    \toprule
    \textbf{Attack} & \textbf{Dataset} & \textbf{UASR}    & \textbf{LASR}    & \textbf{MR}  \\
    \cmidrule(r){1-1}
    \cmidrule(r){2-2}
    \cmidrule(r){3-5}
    \multirow{4}{*}{PWWS} & ChnSetiCorp & 64.87       & 28.06       & 35.09      \\
     & DMSC & 63.20       & 27.27       & 39.72      \\
     & THUCNews & 81.08       & 6.95    & 61.63\\
     & OCNLI &43.80 &30.55&19.79\\
    \cmidrule(r){1-1}
    \cmidrule(r){2-2}
    \cmidrule(r){3-5}
    \multirow{4}{*}{Random} & ChnSetiCorp & 41.88       & 3.48       & 52.74      \\
     & DMSC & 51.51       & 7.03       & 45.53 \\
      & THUCNews & 79.86       & 1.22       & 64.72 \\ 
      & OCNLI & 27.81 & 8.93 & 39.42\\
    \bottomrule
    \end{tabular}%
    }
    \caption{Model performance against unknown attacks. Target models are BERT\textsubscript{base} and training sets are augmented with adversarial texts generated by TextBugger.}
    \label{tab:unknown-attack}
\end{table}
\begin{table}[t]
    \small
    \centering
    \resizebox{0.85\linewidth}{!}{%
    \begin{tabular}{lcccc}
    \toprule
    \textbf{Model}     & \textbf{Acc.} & \textbf{UASR} & \textbf{LASR} & \textbf{MR} \\
    \cmidrule(r){1-1}
    \cmidrule(r){2-5}
    BERT\textsubscript{base}               & 93.02         &78.75	        &60.19	        &13.79            \\
    +graph      & 93.05             & 71.53	        &50.32	        &16.40          \\
    +aug.        & 93.85             & 68.01	            & 48.01	                    & 17.56             \\
     +graph+aug. & 94.15             & 73.16	           & 19.36	    &40.69            \\
    +\sysname            & 92.75             & 36.36             & 29.22             & 12.43       \\ 
    \bottomrule
    \end{tabular}
    }
    \caption{Performance of models enhanced by  different modules against TextBugger attack.}
    \label{tab:ablation}
    \end{table}

\textbf{Generalizability.}
It is seen from Table~\ref{tab:robustness} that both the UASR and LASR against each models protected by \sysname have a different degree of decline.
All models are commonly used in Chinese NLP tasks,
indicating that \sysname has good generalizability on other PLMs, and can be applied on different tasks to defend against various attack.

\textbf{Efficiency.}
We assess efficiency of \sysname via comparing the samples used in data augmentation module with the traditional AT by taking the BERT\textsubscript{base} fine-tuned on {\tt ChnSentiCorp} as an example.
For AT, we conduct attack against the target model with the whole training set and collect successful adversarial texts as the augmentation data.
\sysname only uses 25.40\%, 17.22\% and 16.54\% of the training dataset when performing PWWS, TextBugger and Random attacks respectively, and achieves more robust models than  AT.
It proves that the data augmentation in \sysname is efficient and effective.
In addition, it also indicates the key impact of multimodal fusion.
The complete results are shown in Appendix~\ref{sec:ap_efficiency}.

\subsection{Ablation Study}
\label{sec:ablation}
Then we will discuss the effects of different modules, i.e., adversarial graph and data augmentation.
To observe the gains on adversarial graph, after obtaining \emph{feature embedding}, we simply concatenate it with pre-trained representation and feed them into the model.
To evaluate the effect of data augmentation, we directly attack the target BERT via TextBugger and collect the the intermediate and adversarial texts by taking the BERT\textsubscript{base} fine-tuned on {\tt DMSC} as an example.
The results are shown in Table~\ref{tab:ablation}.
Observe that either using adversarial graph or data augmentation alone can still decrease the UASR and LASR. 
Then, we utilize both adversarial graph and data augmentation to defend BERT against attack, we can see that compared with using one module alone, UASR rises but LASR drops.
When BERT is protected by \sysname, we can obtain the optimal model in general situation with relative low UASR and LASR.
So the two modules we proposed play important role in strengthening robustness of models. 
In addition, the differences between \emph{+graph+aug} and \emph{+\sysname} demonstrate that the effect of simply concatenating \emph{feature embedding $H_{1}$} and \emph{pre-trained embedding $H_{2}$} is relatively limited, which is used in the original AdvGraph.
And \emph{transformer} in \sysname  makes two kinds of embedding fully interact with each other and helps lower the UASR significantly.

\begin{figure}[htbp]
    \setlength{\abovecaptionskip}{2pt}
    \centering  
    \vspace{-0.3cm}
    \subfigure[]{   
    \begin{minipage}{3cm}
    \centering   
    \includegraphics[width=.99\textwidth]{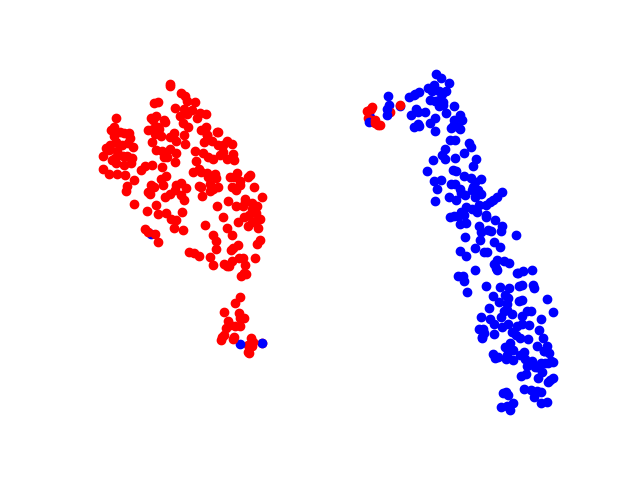} 
    \end{minipage}
    \label{fig:bert_normal}
    }
    \subfigure[]{
    \begin{minipage}{3cm}
    \centering   
    \includegraphics[width=.99\textwidth]{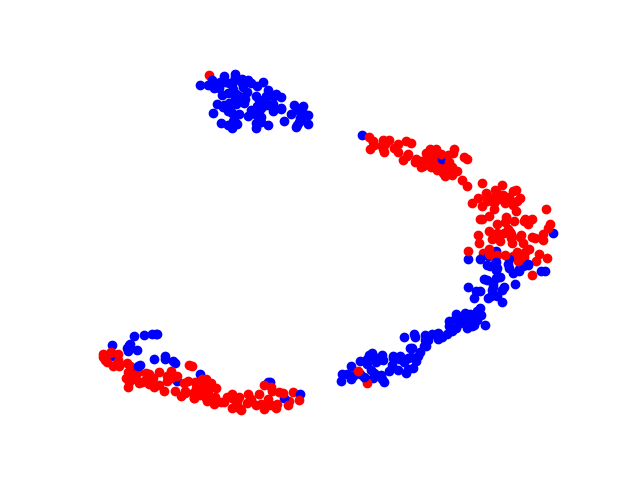}
    \end{minipage}
    \label{fig:bert_adv}
    }
    \subfigure[]{   
    \begin{minipage}{3cm}
    \centering    
    \includegraphics[width=.99\textwidth]{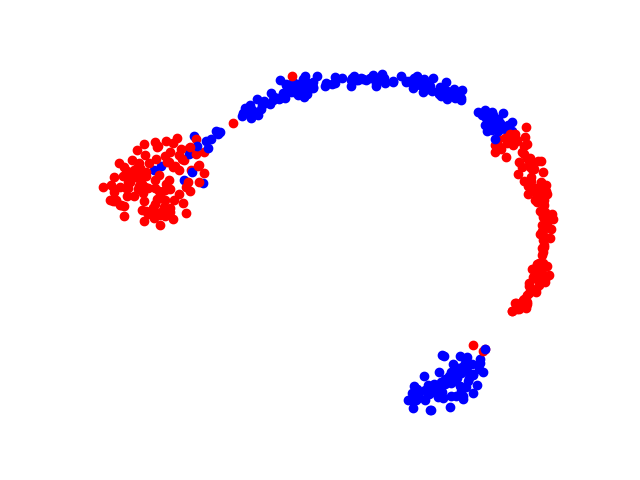}  
    \end{minipage}
    \label{fig:chinesebert_adv}
    }
    \subfigure[]{ 
    \begin{minipage}{3cm}
    \centering  
    \includegraphics[width=.99\textwidth]{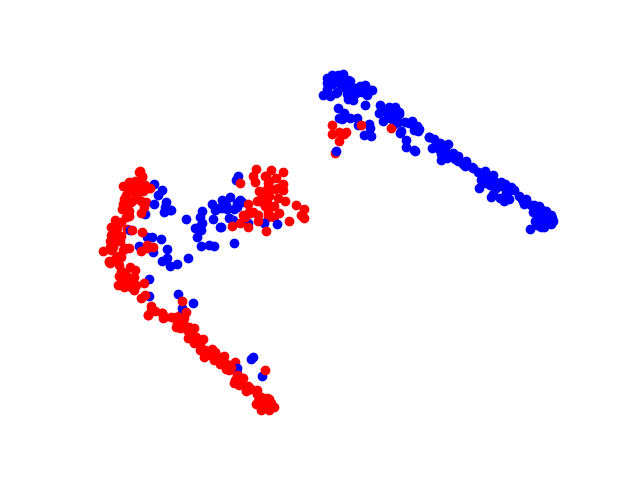}
    \end{minipage}
    \label{fig:robert_adv}
    }
    \caption{Representation Visualization. (a) is benign texts on BERT\textsubscript{base}. (b) is adversarial texts on BERT\textsubscript{base}. (c) is adversarial texts on ChineseBERT. (d) is adversarial texts on BERT\textsubscript{base}+\sysname.} 
    \label{fig:representation} 
    \end{figure}
    
\subsection{Representation Visualization}
To further explore the robustness of different models, we project  representation vectors to two dimensional vectors through T-SNE.
Taking BERT\textsubscript{base} on {\tt DMSC} against TextBugger as an instance, we choose the same 500 samples.
In Figure~\ref{fig:bert_normal}, the decision boundary for benign texts are quite clear.
However in Figure~\ref{fig:bert_adv}, with the addition of adversarial texts, the representations are entangled together, making it difficult to distinguish between the two kinds of data.
As shown in Figure~\ref{fig:chinesebert_adv}, ChineseBERT does not significantly improve this situation.
After being protected by \sysname, most of the adversarial texts are effectively separated in Figure~\ref{fig:robert_adv}, which proves that \sysname exhibits impressive robustness against attacks.

\section{Conclusion}
In this paper, we propose \sysname, a lightweight defense method to strengthen robustness for pre-trained Chinese BERT-based models.
First, we update the current adversarial graph and learn the feature representation of characters based on Chinese pronunciation and glyph.
Then \sysname fuse semantic and feature embeddings to fine-tune the target model with specified downstream NLP task, e.g., classification.
To enhance fusion, we further introduce an augmentation method inspired by curriculum learning.
We evaluate \sysname against several attack algorithms, the results show that \sysname greatly strengthen the model robustness without compromising their normal generalization.
In addition, \sysname can be used to improve most Chinese BERT-based models and it is computationally efficient.

\section*{Limitations}
\sysname\xspace focuses on solving the perturbations that substitutes target character with the ones sharing similar pronunciation or glyph.
In fact, there are other perturbations such as substituting with synonyms, splitting Chinese characters and inserting special characters.
In most of the Chinese BERT models, characters are often seen as the basic unit of operations. So compared with word-based models, e.g., TextCNN and LSTM, it will be harder for adversaries to attack BERT models successfully only by synonyms substitution. Besides, such kind of perturbations can be easily defended by adversarial training.
And the other perturbations are limited, they can be filtered by rules easily.
For completeness of the work, we will include these perturbations into the experiments in future work.
In addition, there are not as many types of Chinese datasets as English ones, and the quality is uneven. We are temporarily unable to verify \sysname on more tasks. In the future, we will consider applying our work on other tasks.
\section*{Ethics Statement}
In our experiments, all adversarial texts are generated from the public benchmark datasets via open-source adversarial attack algorithms.
These generated adversarial texts are used only for the purpose of enhancing the robustness of target models and will not be used for any illegal purposes or unsuitable intentions.
We will also open source the code of our \sysname later to aid those users and researchers involved in pre-trained language models in improving the robustness of their models and thus mitigate the potential physical threats brought by adversarial attacks.

\section*{Acknowledgements}
This work was supported, in part, by the National Key Research and Development Program of China (2021YFB2701000), National Natural Science Foundation of China (61972224), Beijing National Research Center for Information Science and Technology (BNRist) under Grant BNR2022RC01006, Shining Lab and Alibaba Group through Alibaba Research Intern Program.

\bibliographystyle{acl_natbib}
\bibliography{reference}

\clearpage
\appendix

\section{Appendix}
\label{sec:appendix}

\begin{table*}[htbp]
\small
\centering
\begin{tabular}{lcccccc}
\toprule
\textbf{Dataset} & \textbf{Task}  & \textbf{Label} & \textbf{Train} & \textbf{Dev} & \textbf{Test} & \textbf{Length} \\ 
\cmidrule(r){1-1}
\cmidrule(r){2-7}
ChnSentiCorp      & Sentiment Analysis & 2                         & 9,600          & 1,200        & 1,200         & 108                     \\
DMSC             &  Sentiment Analysis & 2                         & 16,000         & 2,000        & 2,000         & 64                      \\
THUCNews         & Text Classification & 5                         & 12,000         & 1,500        & 1,500         & 189                     \\
OCNLI            & Natural Language Inference      & 3                         & 9,985          & 2,950        & -             & 24/11                   \\ 
\bottomrule
\end{tabular}
\caption{Overview of datasets.}
\label{tab:datasets}
\end{table*}

\subsection{Details of Datasets}
\label{sec:ap_dataset}
The datasets we used to evaluate \sysname are shown in Table~\ref{tab:datasets}.
\begin{itemize}
\item \textbf{ChnSentiCorp}: a binary sentiment classification dataset containing 9,600/1,200/1,200 texts in train/dev/test datasets respectively. The content of these texts is the reviews of product and hotel. And their labels are positive or negative.
\item \textbf{DMSC}: a Chinese movie review datasets released by Kaggle. We follow~\citet{li2021enhancing} to sample the original datasets. In our datasets, there are 16,000/2,000/2,000 texts with positive or negative labels in train/dev/test datasets.
\item \textbf{THUCNews}: a news multi-classification dataset with a total of fourteen categories. Like~\cite{zhang2020argot}, we select five of them for experiments. In our dataset, there are 12,000/1,500/1,500 texts in train/dev/test datasets.
\item \textbf{OCNLI}: a Chinese natural language inference dataset from CLUE Benchmark~\cite{xu2020clue}. Each sample contains two sentences, premise and hypothesis. 
We count the average length of the texts separately for these two parts.
And the label (Entailment, Contradiction, or Neutral) represents the relation between premise and hypothesis.
There are 9,985 and 2,950 samples in train and dev datasets. As the samples in test datasets don't have labels, we use the dev datasets to conduct robustness evaluation.
\end{itemize}

\subsection{Details of Attack algorithms}
\label{sec:ap_attack}
we utilize three widely used attacks, i.e., PWWS~\cite{ren2019generating}, TextBugger~\cite{li2018textbugger} and random attack in the black-box setting to evaluate the robustness of \sysname. 
In order to adapt the attack algorithms to Chinese text, we have made some modifications to them.
\begin{itemize}
\item \textbf{PWWS}: It first generates a candidate set for each word, including the synonym of the word and a similar name entity, and then replaces the words to find the optimal candidate word according to the confidence drop.
The next step is to determine which word in the text should be replaced preferentially. 
It still uses confidence drop to evaluate the priority.
In order to make PWWS generate Chinese adversarial texts, we changed the candidate words from the original synonyms and name entities to phrases composed of Chinese phonetic and glyph-similar words.
\item \textbf{TextBugger}: It first segments the text into individual sentences, and it ranks the importance of the sentences by querying the target model with single sentence in turn. Then the words in the sentence are removed one by one, and the drops of confidence are used to determine the importance of the words. In order to make TextBugger generate Chinese adversarial texts, we only keep one of its bug generation methods, \emph{Word Substitution}, and the candidate words are phrases composed of Chinese phonetic and glyph-similar words.
\item \textbf{Random Attack}: It is the baseline method compared with other two attack algorithms. It randomly selects words in the text and use the candidate words to substitute the original one.
\end{itemize}
In general, we use Chinese characters with similar glyph or pronunciation to form candidate words (there are at most 40 candidate words). And after locating the important parts through the corresponding attack algorithm, we select suitable candidate words to replace the original words.
Specifically, we use the same perturbation strategies in both data augmentation and robustness evaluations. 

\subsection{Details of $epsilon_{max}$}
\label{sec:ap_epsilon}
The details of $\epsilon_{max}$ we used in experiment are shown in Table~\ref{tab:epsilon}.

\begin{table}[t]
\centering
\small
\begin{tabular}{cccc}
\toprule
\multirow{2}{*}{\textbf{Dataset}} & \multicolumn{3}{c}{\textbf{Attack Algorithm}}         \\ 
 & \textbf{PWWS} & \textbf{TextBugger} & \textbf{Random} \\ 
\cmidrule(r){1-1}
\cmidrule(r){2-4}
ChnSentiCorp        & 0.45          & 0.45                & 0.45            \\
DMSC                & 0.45          & 0.45                & 0.45            \\
THUCNews            & 0.3           & 0.3                 & 0.3             \\
OCNLI               & 0.3           & 0.3                 & 0.1             \\ \bottomrule
\end{tabular}
\caption{The $\epsilon_{max}$ we used in experiment.}

\label{tab:epsilon}
\end{table}

\subsection{Results of Attacking Hypothesis in OCNLI}
\label{sec:ap_hypothesis}
\subsubsection{Model Performance}
The results are presented in Table~\ref{tab:clean-perf-hypo}.
We can see that ChineseBERT outperforms other models on OCNLI.
The data augmentation does affects the accuracy in some degree, especially the models using hypothesis examples generated by PWWS and TextBugger algorithms.
But in some cases, models defended by \sysname can help models improve accuracy on benign texts, which is consistent with the results of other tasks and datasets.
In addition, SC reduces the accuracy of models due to its own errors. 

\begin{table}[t]
    \centering
    \small
    
    \begin{tabular}{cc}
    \toprule
    \textbf{Model}     &      \textbf{OCNLI} \\ 
    \cmidrule(r){1-1}
    \cmidrule(r){2-2}
    ChineseBERT              &\textbf{73.20}   \\ 
    \cmidrule(r){1-1}
    \cmidrule(r){2-2}
    BERT\textsubscript{base}       &71.57    \\
    +SC       &70.57      \\
    +\sysname (PWWS)   &  68.34   \\
    +\sysname (TextBugger)      & 68.48    \\
    +\sysname (Random)    &70.52   \\ 
    \cmidrule(r){1-1}
    \cmidrule(r){2-2}
    BERT\textsubscript{wwm}     &70.33         \\
    +SC    & 69.08      \\
    +\sysname (PWWS)       &69.41        \\
    +\sysname (TextBugger)     &67.87         \\
    +\sysname (Random)      &70.30         \\ 
    \cmidrule(r){1-1}
    \cmidrule(r){2-2}
    BERT\textsubscript{wwm/ext}    &71.16          \\
    +SC       &70.68  \\
    +\sysname (PWWS)        &69.29\\
    +\sysname (TextBugger)    &  71.27      \\
    +\sysname (Random)      &71.56        \\ 
    \cmidrule(r){1-1}
    \cmidrule(r){2-2}
    RoBERTa\textsubscript{wwm/ext}   &71.29        \\
    +SC          &70.88   \\
    +\sysname (PWWS)       & 70.07        \\
    +\sysname (TextBugger)      & 70.30        \\
    +\sysname (Random)      & 71.61      \\ 
    \bottomrule
    \end{tabular}
        \caption{Model performances on benign texts.}

    \label{tab:clean-perf-hypo}
    \vspace{-0.3cm}
    \end{table}

\subsubsection{Robustness Against Attack}
The results of attacking hypothesis in OCNLI are shown in Table~\ref{tab:robustness-hypo}.
It proves that \sysname can enhance the robustness of BERT-based models.
We can see that \sysname is not as effective in attacking hypothesis as it is on other tasks and datasets. This may be because the length of the hypothesis sentences is very short, and modifying a small number of characters will have a great impact on the original embedding, so it is easy to attack successfully.

\begin{table*}[htbp]

    \centering
    \small
    \begin{tabular}{cccccccccc}
    \toprule
    \multicolumn{1}{c}{\multirow{2}{*}{\textbf{Model}}} & \multicolumn{3}{c}{\textbf{PWWS}}                                  & \multicolumn{3}{c}{\textbf{TextBugger}}                            & \multicolumn{3}{c}{\textbf{Random}}            \\
    \cmidrule(r){2-4}
    \cmidrule(r){5-7}
    \cmidrule(r){8-10}
    \multicolumn{1}{c}{}                                & \textbf{UASR} & \textbf{LASR} & \multicolumn{1}{c}{\textbf{Modi.}} & \textbf{UASR} & \textbf{LASR} & \multicolumn{1}{c}{\textbf{Modi.}} & \textbf{UASR} & \textbf{LASR} & \textbf{Modi.} \\ 
    \cmidrule(r){1-1}
    \cmidrule(r){2-10}
    \multicolumn{10}{c}{{\tt OCNLI}}                                                                                                                                                                                                                         \\ \cmidrule(r){1-1}
    \cmidrule(r){2-10}
    \multicolumn{1}{c}{ChineseBERT}                     & 99.05        & 54.86         & \multicolumn{1}{c}{21.23}          & 99.86         & 48.65         & \multicolumn{1}{c}{22.72}          & 79.19         & 10.27          & 48.84          \\ 
    \cmidrule(r){1-1}
    \cmidrule(r){2-4}
    \cmidrule(r){5-7}
    \cmidrule(r){8-10}
    \multicolumn{1}{c}{BERT\textsubscript{base}}               & \textbf{94.77}         & 52.34         & \multicolumn{1}{c}{20.64}          & 98.76         & 47.52         & \multicolumn{1}{c}{22.71}          & 70.80         & 11.98          & 41.21          \\
    \multicolumn{1}{c}{+SC}               & 95.24         & 53.15         & \multicolumn{1}{c}{20.10}          & 98.32        & 48.53         & \multicolumn{1}{c}{22.25}          & 71.61         & 10.07          & 43.10          \\
    \multicolumn{1}{c}{+\sysname}                          & 97.83        & \textbf{47.39}        & \multicolumn{1}{c}{\textbf{24.07}}          & \textbf{97.95}        & \textbf{36.40}        & \multicolumn{1}{c}{\textbf{26.56}}          & \textbf{70.00}       & 10.97        & 46.05          \\ 
    \cmidrule(r){1-1}
    \cmidrule(r){2-4}
    \cmidrule(r){5-7}
    \cmidrule(r){8-10}
    \multicolumn{1}{c}{BERT\textsubscript{wwm}}                & 96.34	    & 51.41	    & \multicolumn{1}{c}{21.26}              & 99.44	        & 45.92           & \multicolumn{1}{c}{23.13}         & 75.35	       & \textbf{7.75}	        & 48.95             \\
    \multicolumn{1}{c}{+SC}               & 97.30         & 55.62         & \multicolumn{1}{c}{20.33}          & 99.72         & 49.36         & \multicolumn{1}{c}{22.47}          & 76.81         & 11.81          & 46.00          \\
    \multicolumn{1}{c}{+\sysname}    & 97.99             & 49.43             & \multicolumn{1}{c}{22.76}      & 99.86      & 40.11            & \multicolumn{1}{c}{25.64}              & 73.45             & 10.25            & 47.99              \\ 
    \cmidrule(r){1-1}
    \cmidrule(r){2-4}
    \cmidrule(r){5-7}
    \cmidrule(r){8-10}
    \multicolumn{1}{c}{BERT\textsubscript{wwm/ext}}            & 97.36	        & 51.11             & \multicolumn{1}{c}{22.09}              & 99.86	        & 45.14	        & \multicolumn{1}{c}{23.44}              & 78.47	            & 8.75	        & 50.52             \\
    \multicolumn{1}{c}{+SC}               & 97.07         & 54.95         & \multicolumn{1}{c}{21.06}          & 99.58         & 49.09         & \multicolumn{1}{c}{21.73}          & 79.78         & 8.65          & 48.14          \\
    \multicolumn{1}{c}{+\sysname}                          & 96.25            & 49.64             & \multicolumn{1}{c}{21.36}              & 99.58             & 40.36             & \multicolumn{1}{c}{25.84}              & 75.21            & 9.10             & \textbf{52.19}             \\ 
    \cmidrule(r){1-1}
    \cmidrule(r){2-4}
    \cmidrule(r){5-7}
    \cmidrule(r){8-10}
    \multicolumn{1}{c}{RoBERTa\textsubscript{wwm/ext}}         & 96.57        & 51.92	          & \multicolumn{1}{c}{21.84}              & 99.18	        & 45.19	             & \multicolumn{1}{c}{23.83}              & 79.26	        & 8.93	& 51.04           \\
    \multicolumn{1}{c}{+SC}               & 96.83         & 52.97         & \multicolumn{1}{c}{21.04}          & 99.17         & 47.17         & \multicolumn{1}{c}{22.34}          & 81.24         & 9.93          & 49.74          \\
    \multicolumn{1}{c}{+\sysname}                   & 96.03     & 52.19          & \multicolumn{1}{c}{22.30}              & 99.17           & 40.33            & \multicolumn{1}{c}{25.77}              &  71.91            & 13.16            & 42.64             \\ 
    \bottomrule
    \end{tabular}
    \caption{Model performance against different attacks.}
    \label{tab:robustness-hypo}
    \end{table*}

\subsection{Representation Visualization}
\begin{figure}[htbp]
    \centering  
    \subfigure[]{
    \begin{minipage}{3cm}
    \centering   
    \includegraphics[width=.95\textwidth]{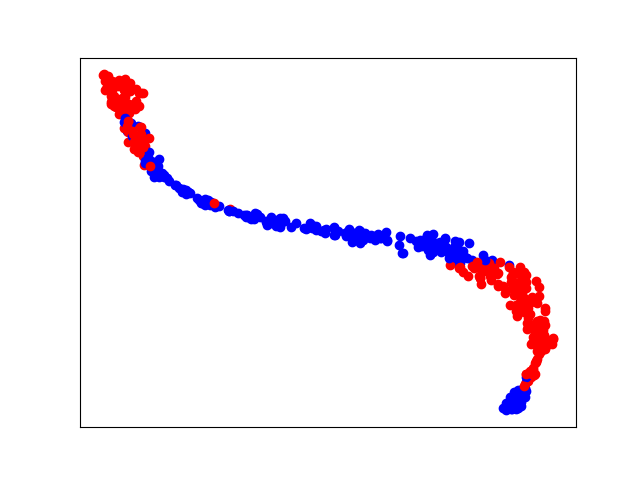}
    \end{minipage}
    \label{fig:bertwwm_adv}
    }
    \subfigure[]{ 
    \begin{minipage}{3cm}
    \centering  
    \includegraphics[width=.95\textwidth]{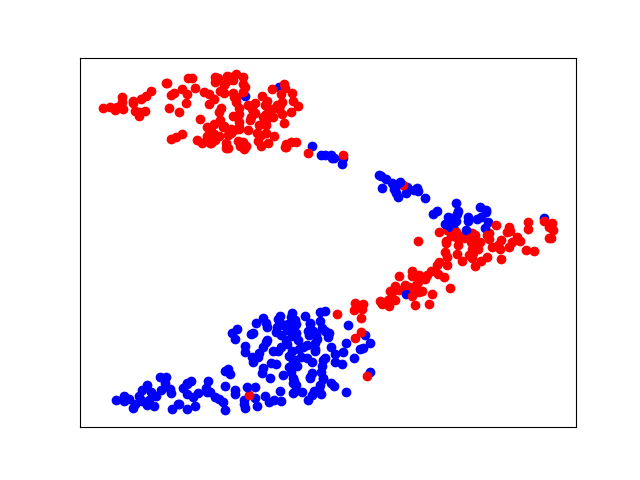}
    \end{minipage}
    \label{fig:bertwwmrobert_adv}
    }
    \subfigure[]{
    \begin{minipage}{3cm}
    \centering   
    \includegraphics[width=.95\textwidth]{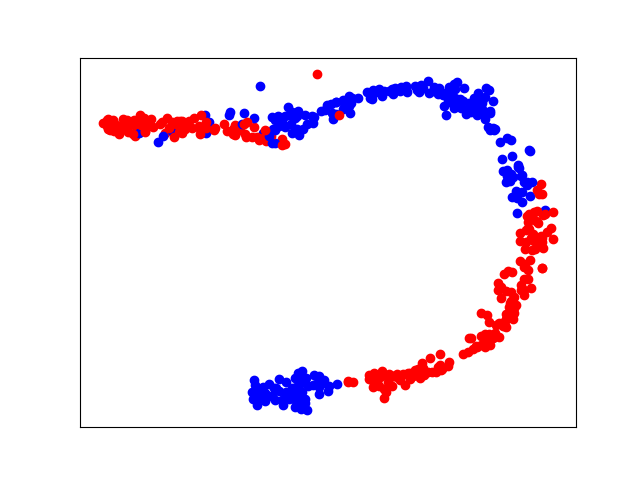}
    \end{minipage}
    \label{fig:bertwwmext_adv}
    }
    \subfigure[]{ 
    \begin{minipage}{3cm}
    \centering  
    \includegraphics[width=.95\textwidth]{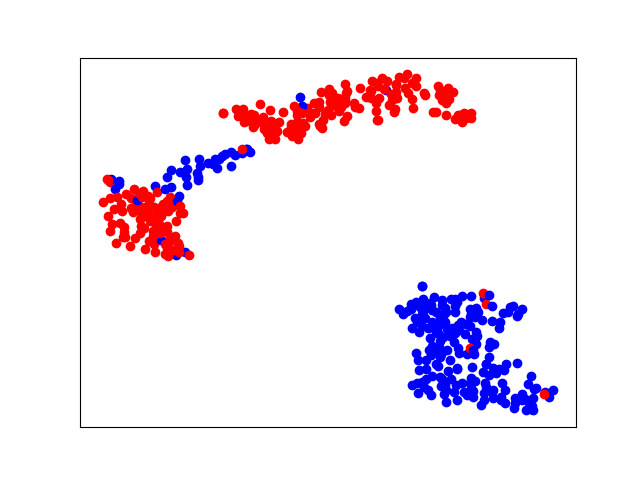}
    \end{minipage}
    \label{fig:bertwwmextrobert_adv}
    }
    \subfigure[]{
    \begin{minipage}{3cm}
    \centering   
    \includegraphics[width=.95\textwidth]{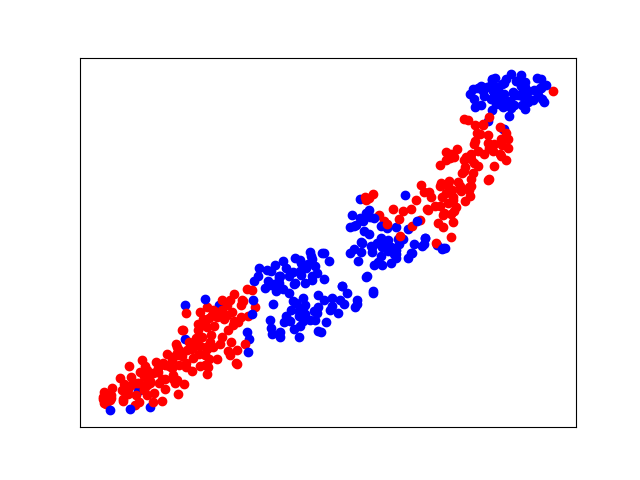}
    \end{minipage}
    \label{fig:roberta_adv}
    }
    \subfigure[]{ 
    \begin{minipage}{3cm}
    \centering  
    \includegraphics[width=.95\textwidth]{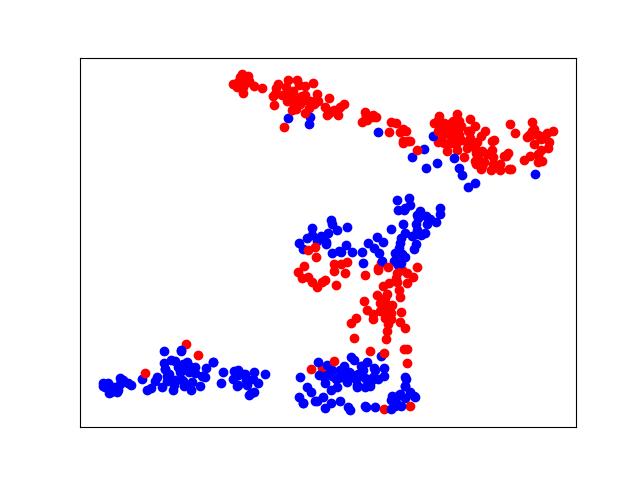}
    \end{minipage}
    \label{fig:robertarobert_adv}
    }
    \caption{Representation with different models. (a) is adversarial texts on BERT\textsubscript{wwm}. (b) is adversarial texts on BERT\textsubscript{wwm}+\sysname. (c) is adversarial texts on BERT\textsubscript{wwm/ext}. (d) is adversarial texts on BERT\textsubscript{wwm/ext}+\sysname. (e) is adversarial texts on RoBERTa\textsubscript{wwm/ext}. (f) is adversarial texts on RoBERTa\textsubscript{wwm/ext}+\sysname.} 
    \label{fig:all_representation} 
    \end{figure}

We add the representation visualization results on other models and their corresponding variant models protected by \sysname in Figure~\ref{fig:all_representation}. 
We still use models trained on {\tt DMSC} dataset against TextBugger attack and choose the same 500 samples. 
Obviously, most of the adversarial texts are effectively separated after being defended by \sysname, which once again proves that \sysname significantly strengthen models' robustness and can be easily exploited on other pre-trained language models.

\begin{figure*}[htbp]
    \centering  
    \subfigure[PWWS Attack on {\tt ChnSentiCorp}]{   
    \begin{minipage}{5cm}
    \centering   
    \includegraphics[width=.95\textwidth]{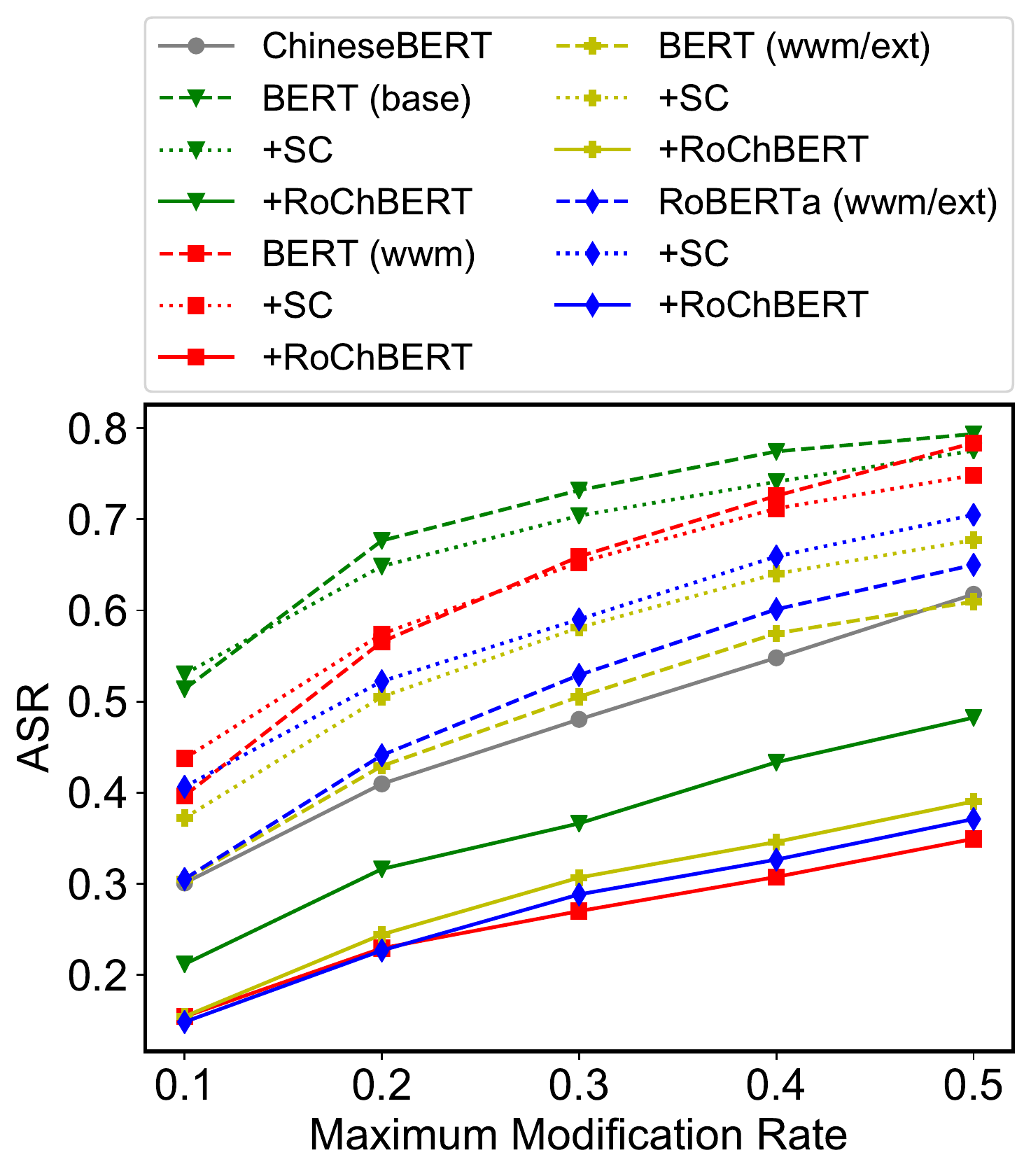} 
    \end{minipage}
    }
    \subfigure[TextBugger Attack on {\tt ChnSentiCorp}]{
    \begin{minipage}{5cm}
    \centering   
    \includegraphics[width=.95\textwidth]{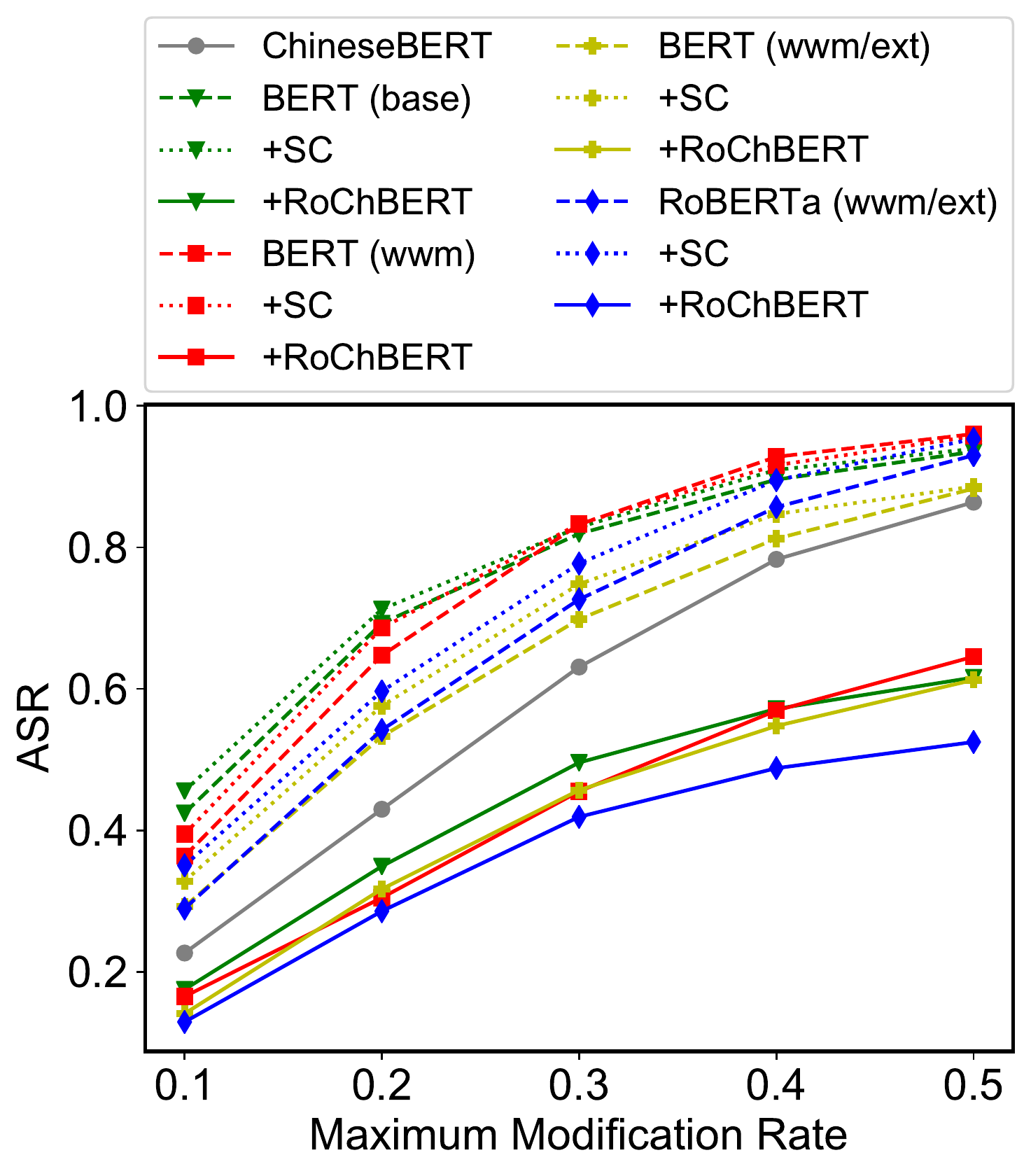}
    \end{minipage}
    }
    \subfigure[Random Attack on {\tt ChnSentiCorp}]{   
    \begin{minipage}{5cm}
    \centering    
    \includegraphics[width=.95\textwidth]{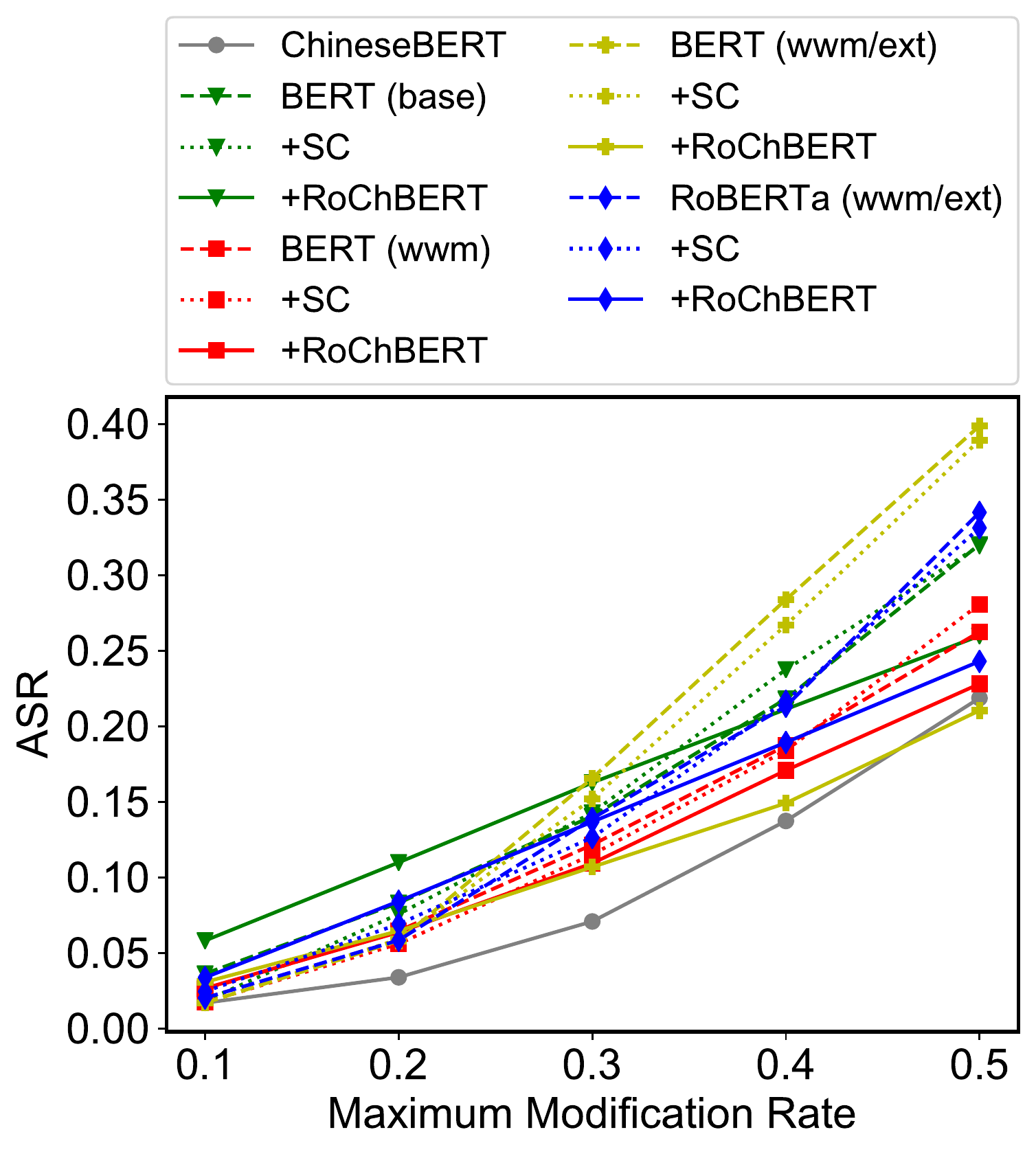}  
    \end{minipage}
    }
    
    \caption{Relation of modification rate and ASR on {\tt ChnSentiCorp}.} 
    \label{fig:relation-chn} 
    \end{figure*}

\begin{figure*}[htbp]
    \centering  
    \subfigure[PWWS Attack on {\tt DMSC}]{   
    \begin{minipage}{5cm}
    \centering   
    \includegraphics[width=.95\textwidth]{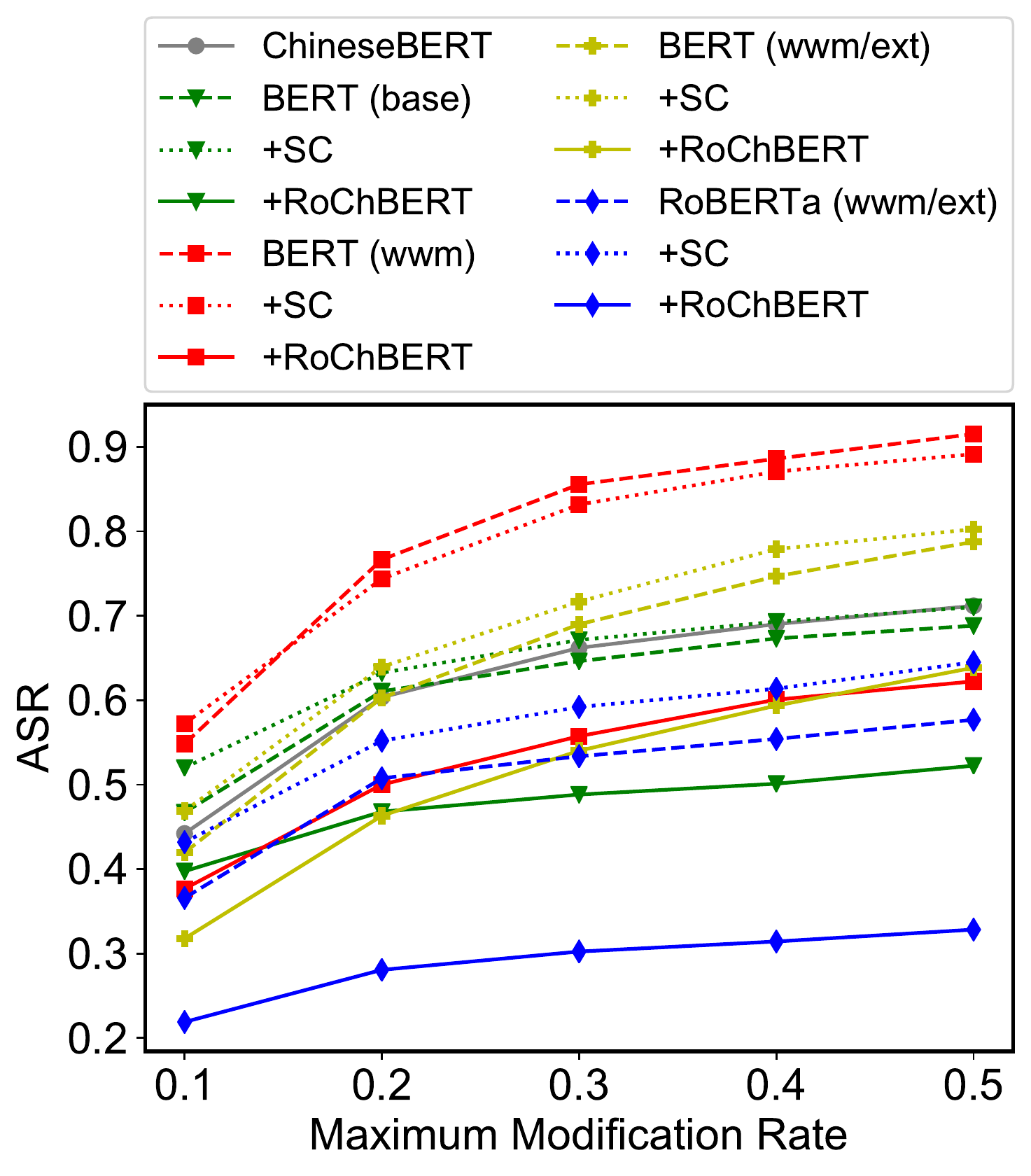} 
    \end{minipage}
    }
    \subfigure[TextBugger Attack on {\tt DMSC}]{
    \begin{minipage}{5cm}
    \centering   
    \includegraphics[width=.95\textwidth]{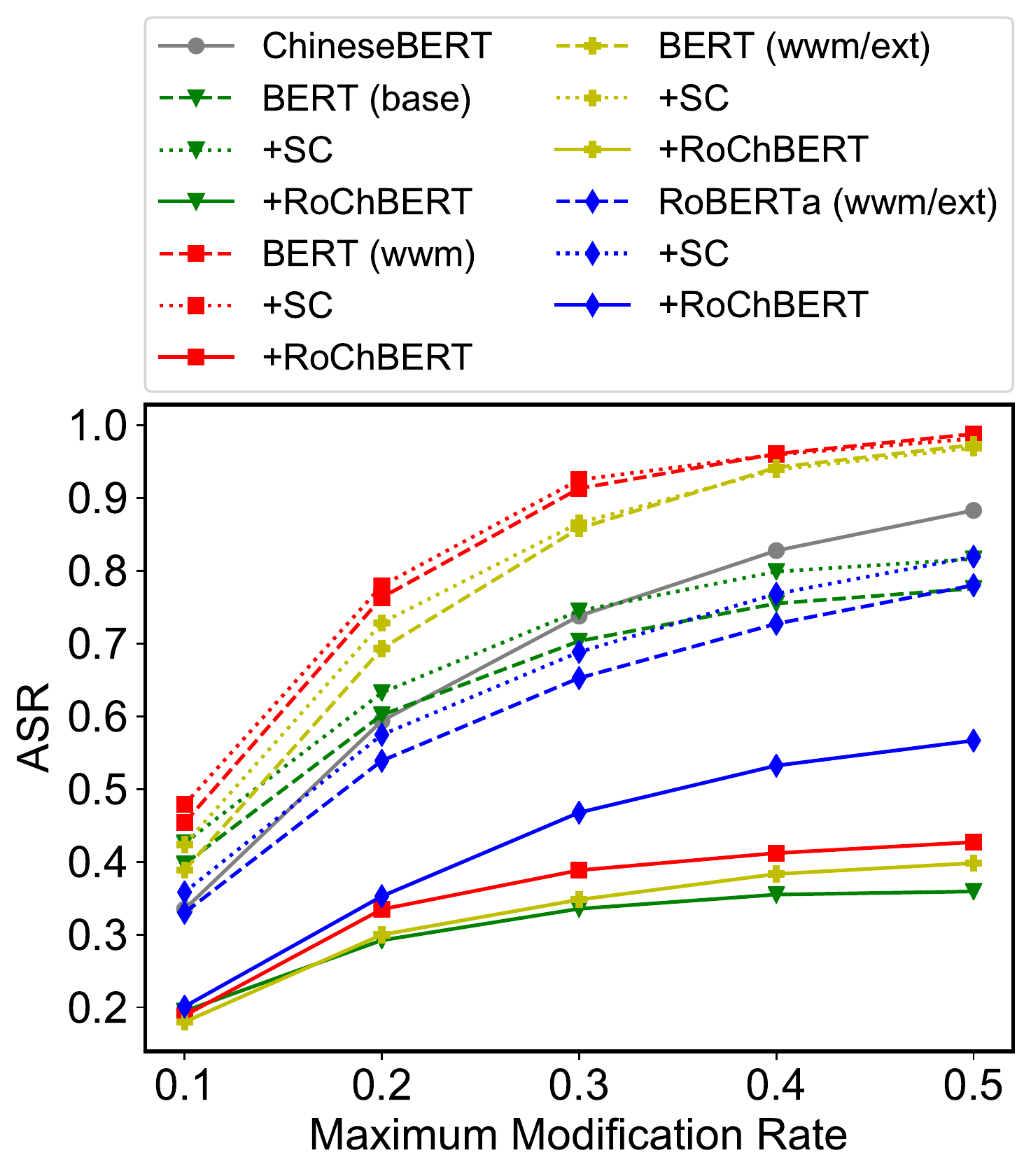}
    \end{minipage}
    }
    \subfigure[Random Attack on {\tt DMSC}]{   
    \begin{minipage}{5cm}
    \centering    
    \includegraphics[width=.95\textwidth]{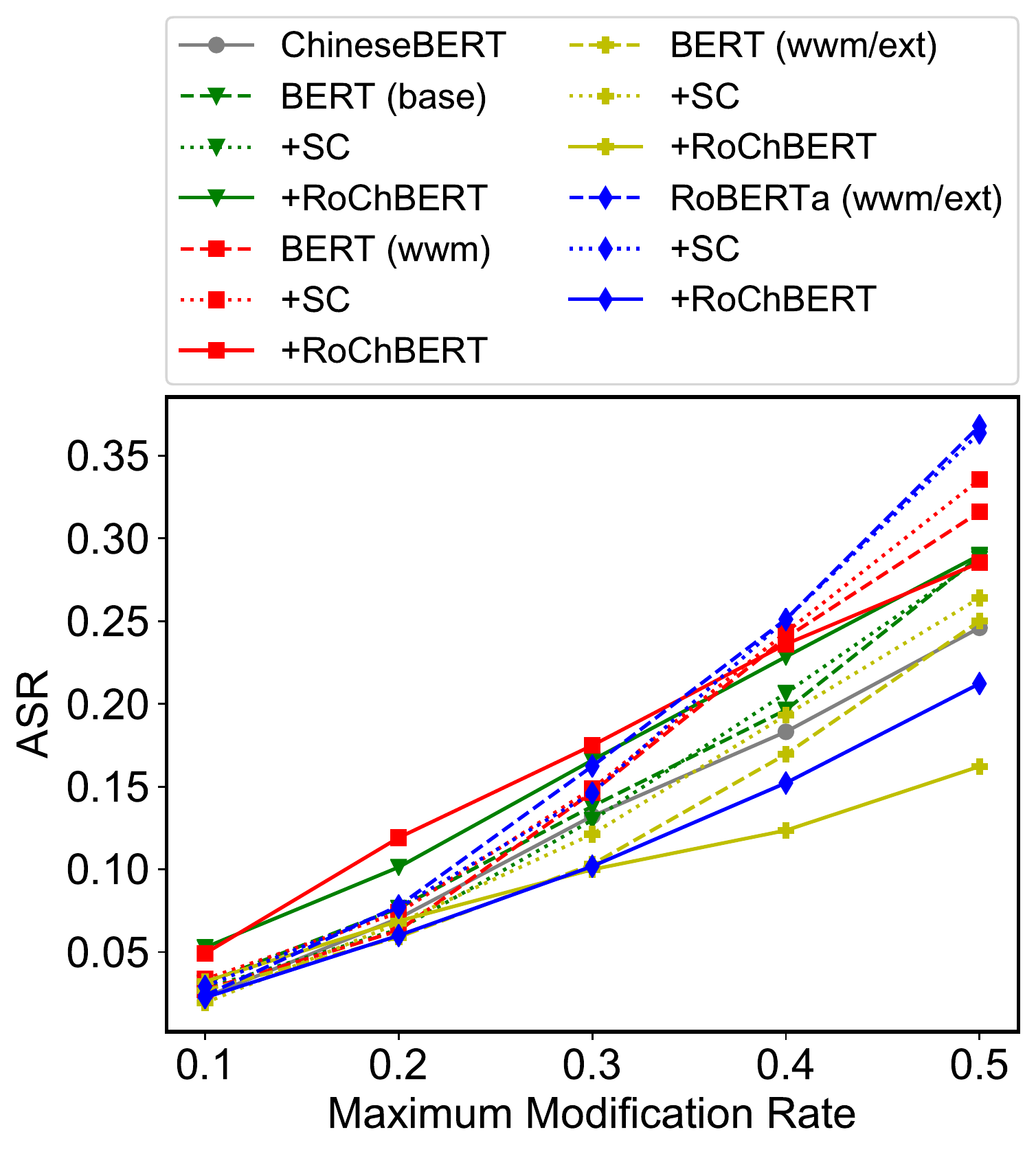}  
    \end{minipage}
    }
    
    \caption{Relation of modification rate and ASR on {\tt DMSC}.} 
    \label{fig:relation-dmsc} 
    \end{figure*}
    
\begin{figure*}[htbp]
    \centering  
    \subfigure[PWWS Attack on {\tt THUCNews}]{   
    \begin{minipage}{5cm}
    \centering   
    \includegraphics[width=.95\textwidth]{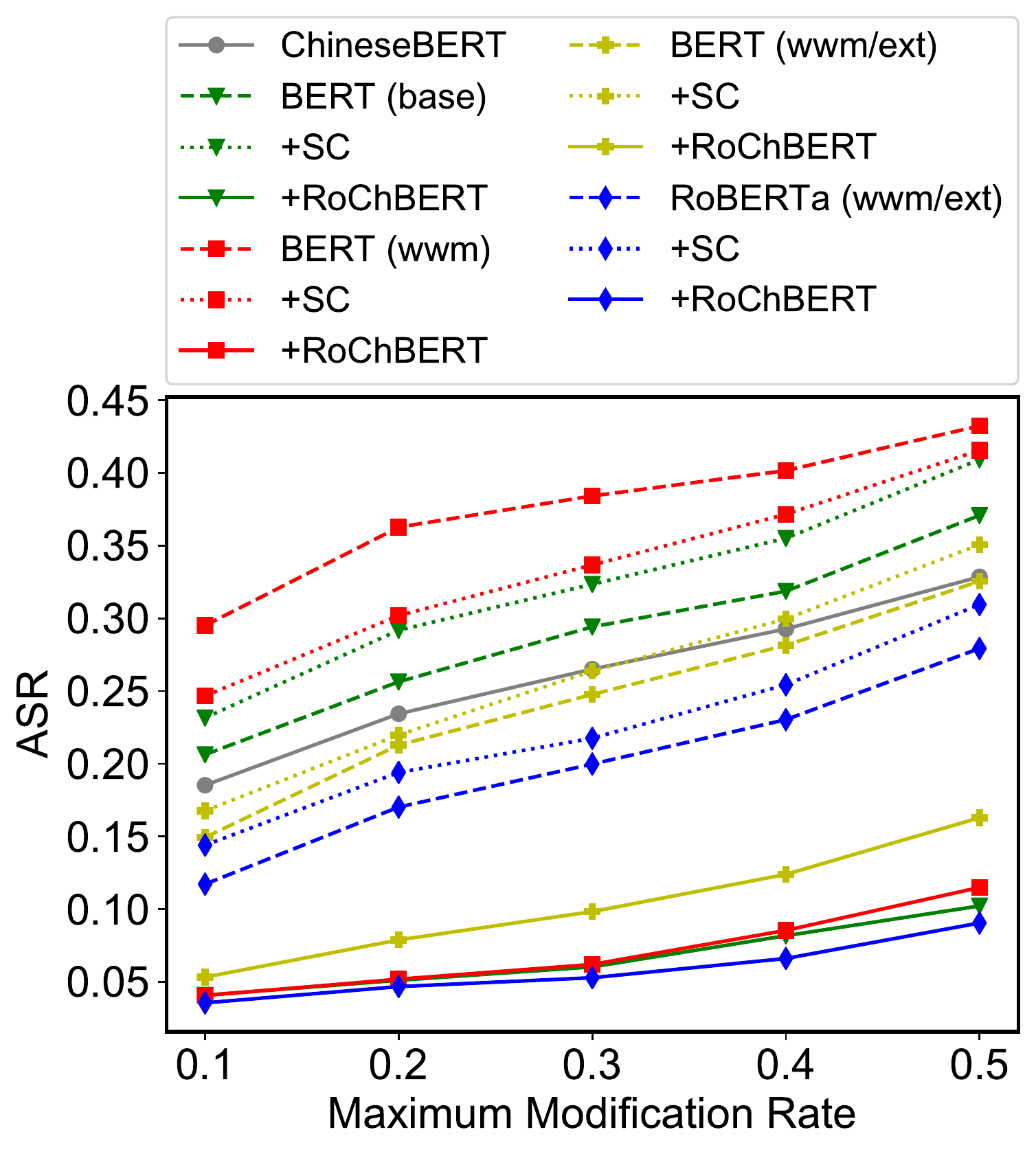} 
    \end{minipage}
    }
    \subfigure[TextBugger Attack on {\tt THUCNews}]{
    \begin{minipage}{5cm}
    \centering   
    \includegraphics[width=.95\textwidth]{figure/thu_textbugger.pdf}
    \end{minipage}
    }
    \subfigure[Random Attack on {\tt THUCNews}]{   
    \begin{minipage}{5cm}
    \centering    
    \includegraphics[width=.95\textwidth]{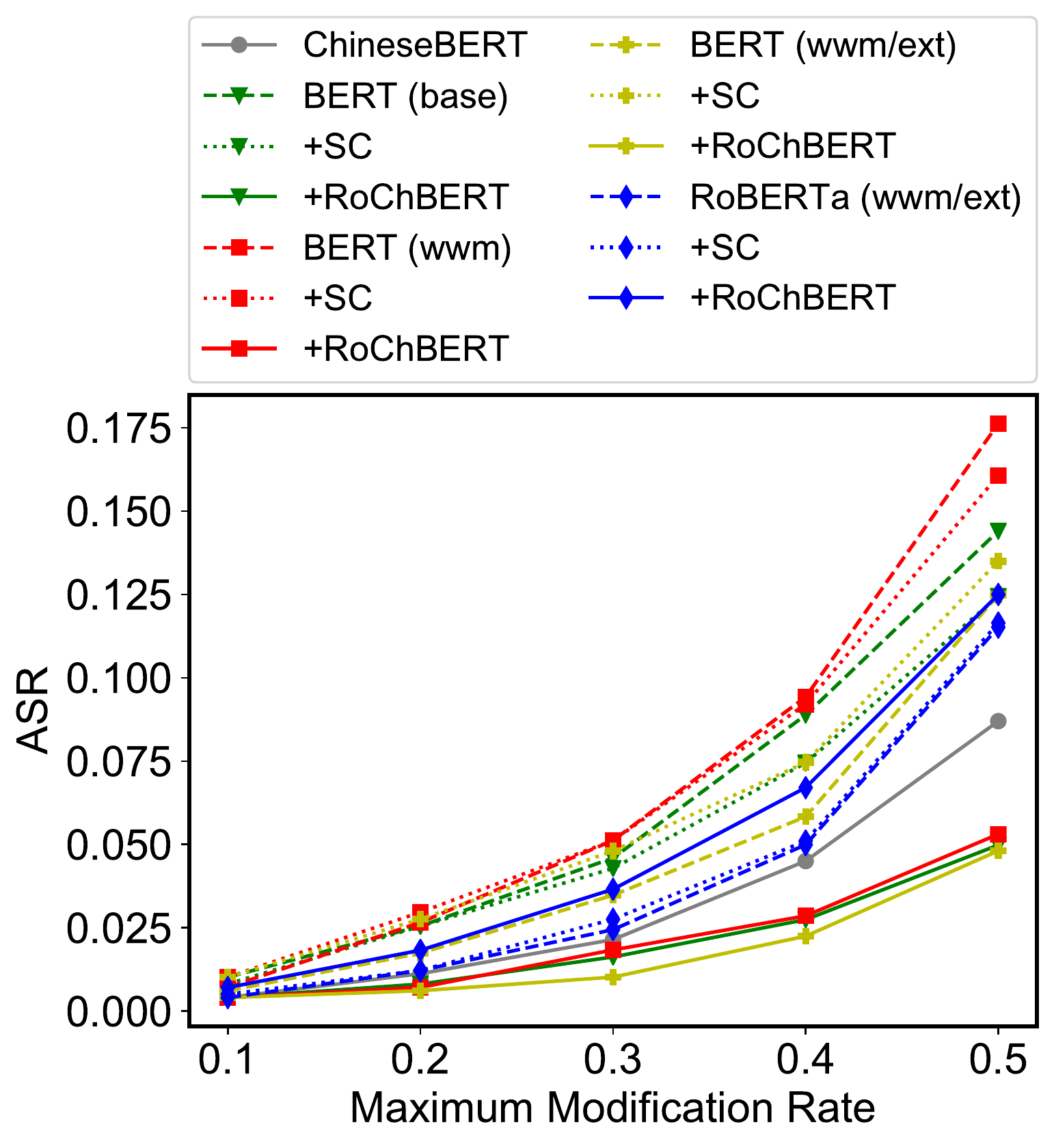}  
    \end{minipage}
    }
    
    \caption{Relation of modification rate and ASR on {\tt THUCNews}.} 
    \label{fig:relation-thu} 
    \end{figure*}
    
\begin{figure*}[htbp]
    \centering  
    \subfigure[PWWS Attack on {\tt OCNLI}-premise]{   
    \begin{minipage}{5cm}
    \centering   
    \includegraphics[width=.95\textwidth]{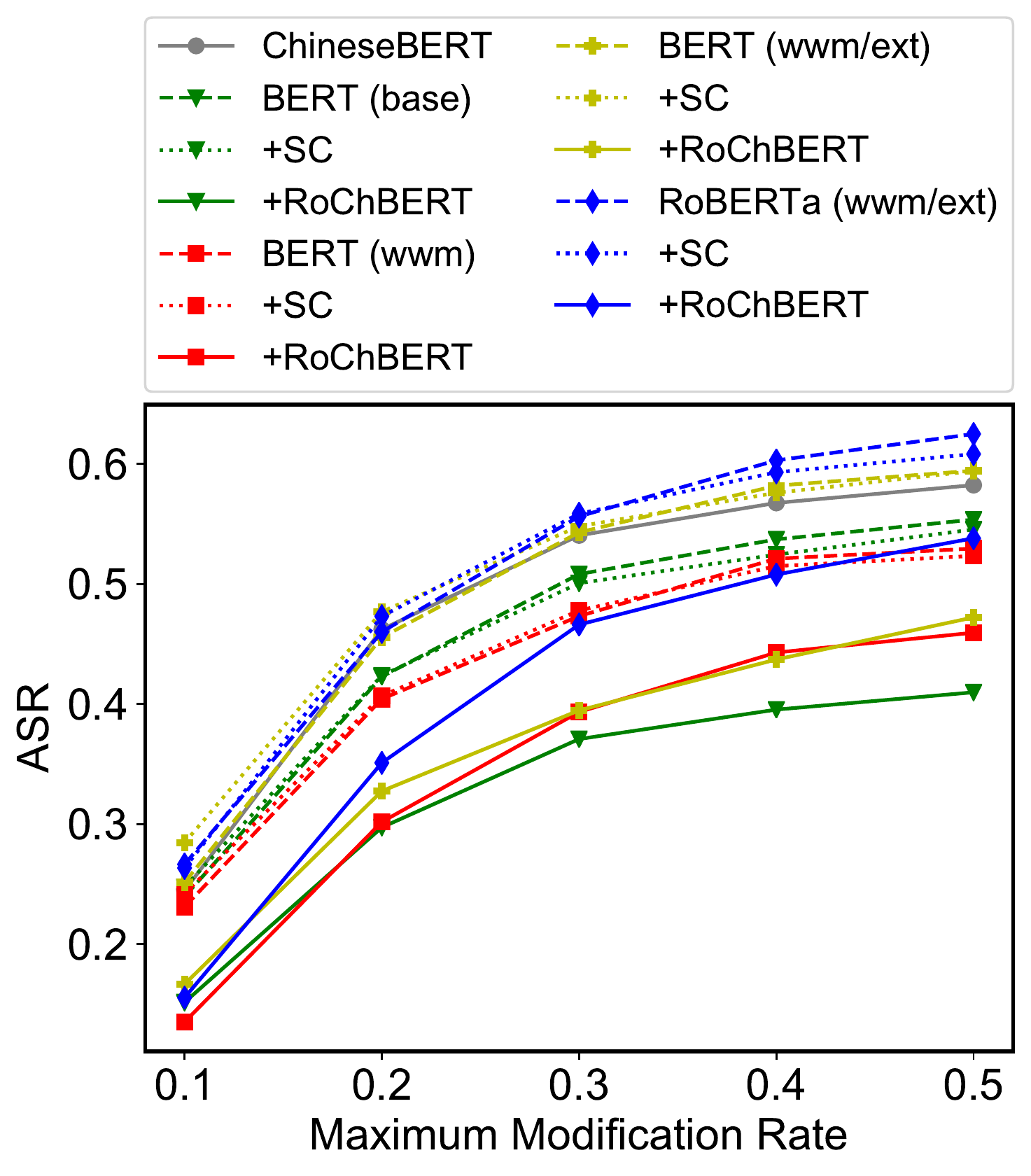} 
    \end{minipage}
    }
    \subfigure[TextBugger Attack on {\tt OCNLI}-premise]{
    \begin{minipage}{5cm}
    \centering   
    \includegraphics[width=.95\textwidth]{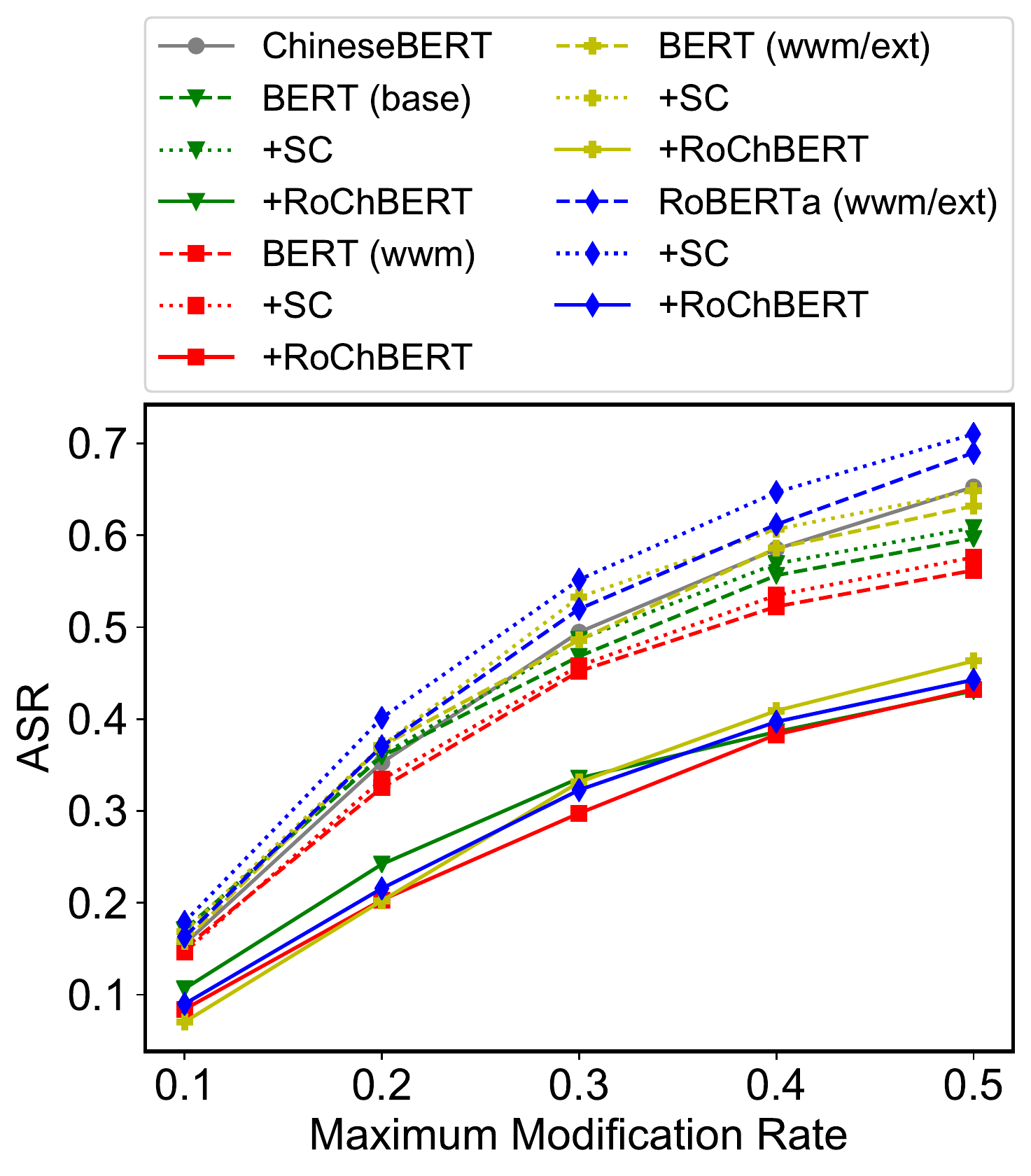}
    \end{minipage}
    }
    \subfigure[Random Attack on {\tt OCNLI}-premise]{   
    \begin{minipage}{5cm}
    \centering    
    \includegraphics[width=.95\textwidth]{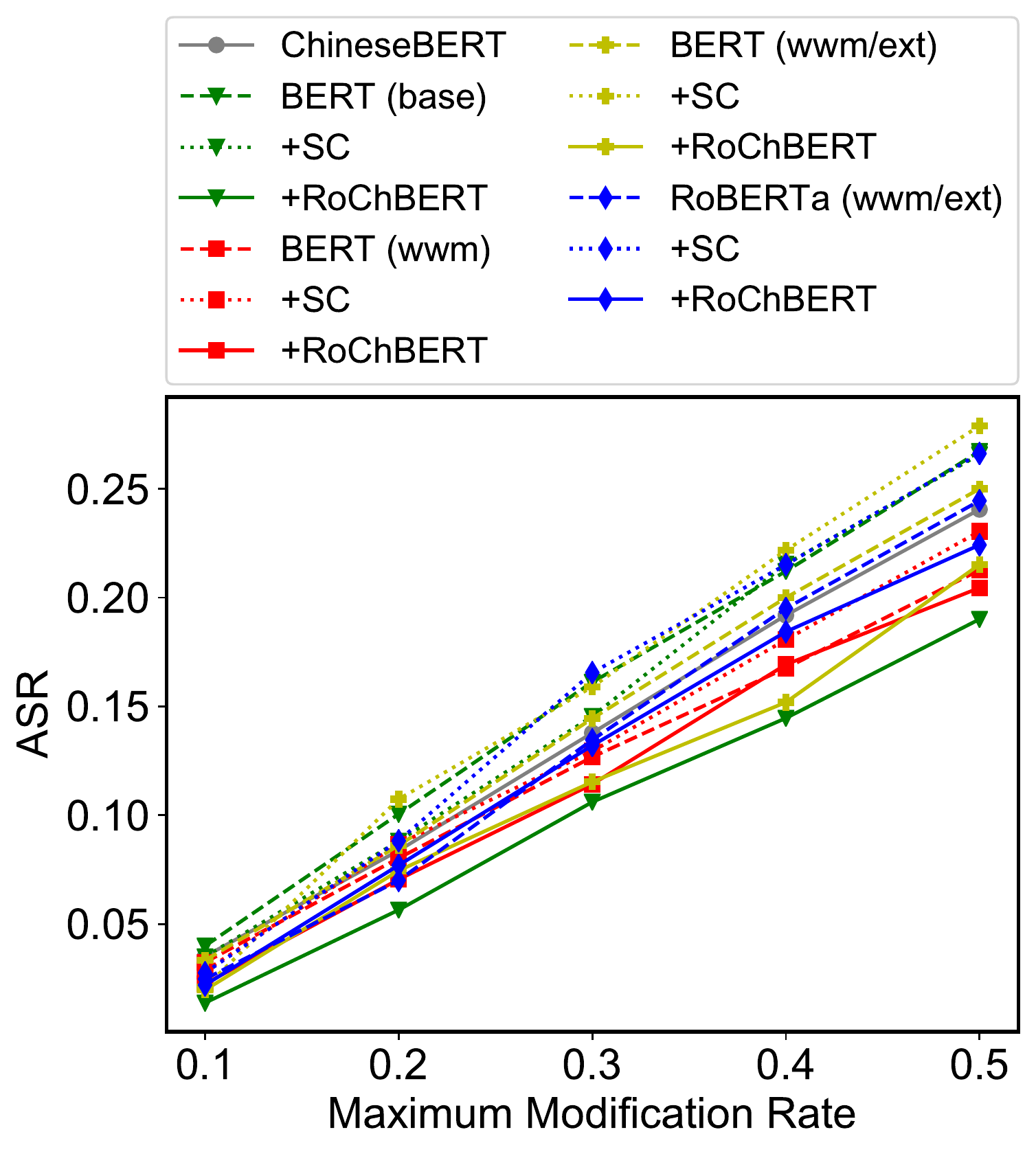}  
    \end{minipage}
    }
    \subfigure[PWWS Attack on {\tt OCNLI}-hypothesis]{   
    \begin{minipage}{5cm}
    \centering   
    \includegraphics[width=.95\textwidth]{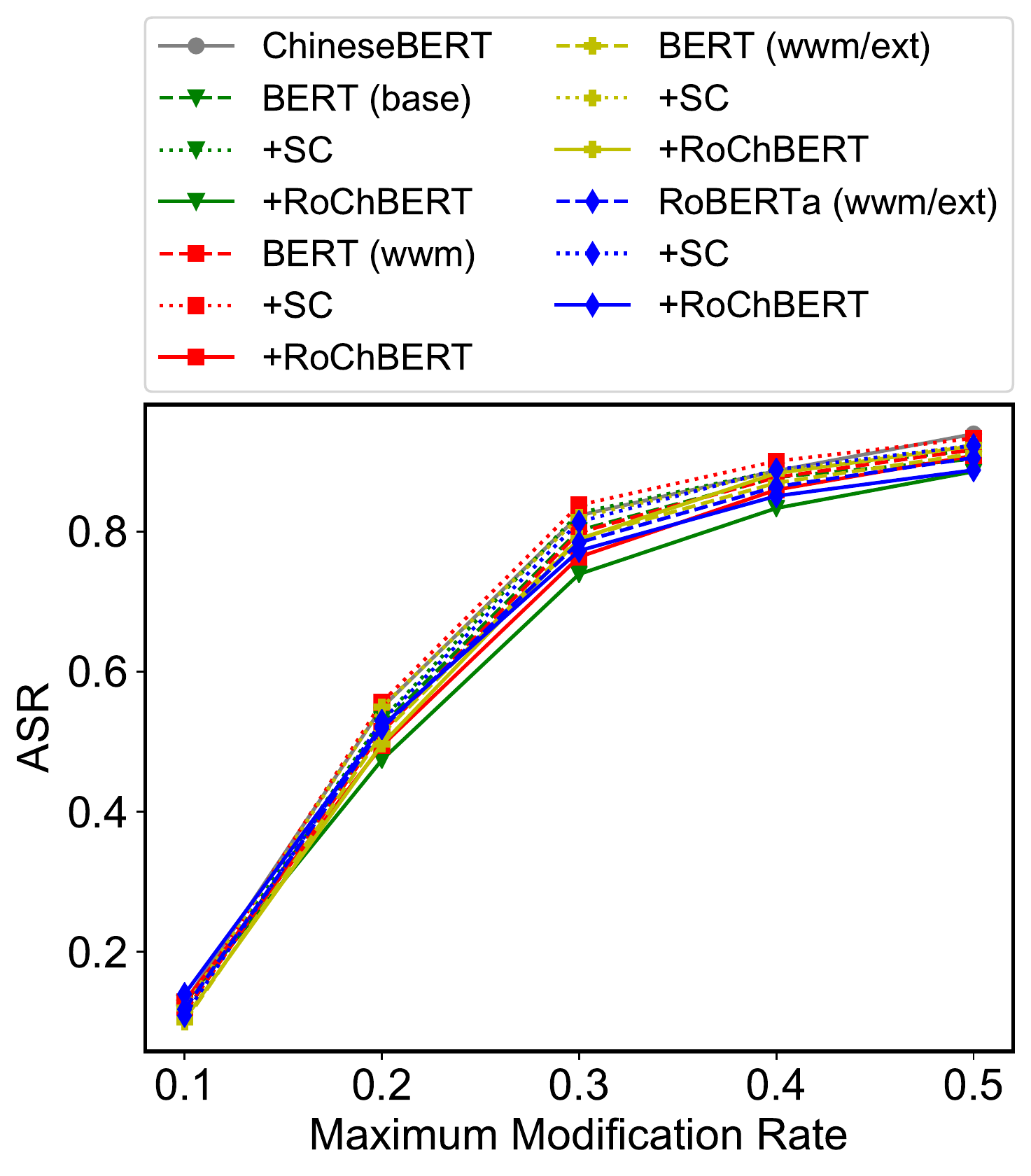} 
    \end{minipage}
    }
    \subfigure[TextBugger Attack on {\tt OCNLI}-hypothesis]{
    \begin{minipage}{5cm}
    \centering   
    \includegraphics[width=.95\textwidth]{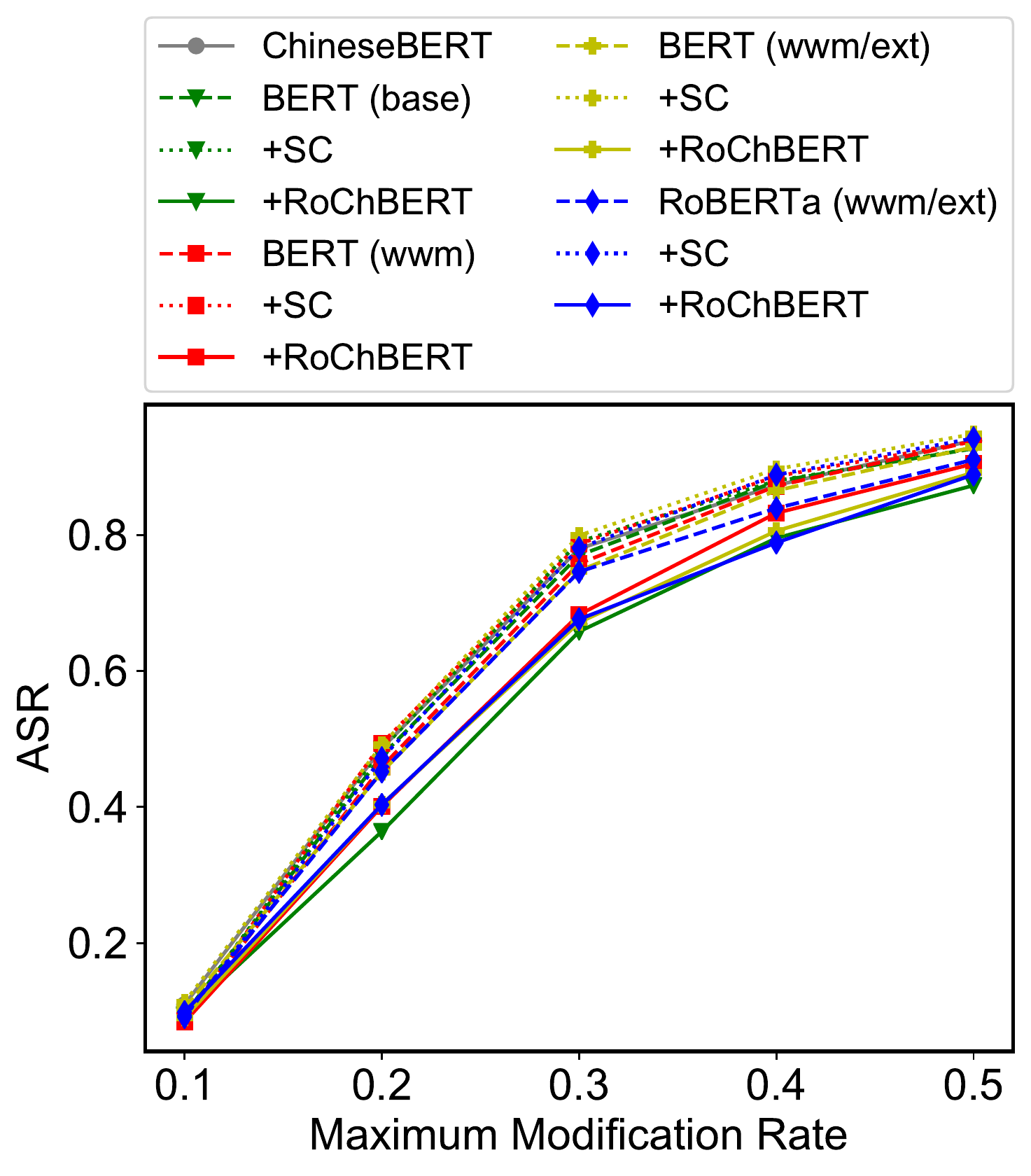}
    \end{minipage}
    }
    \subfigure[Random Attack on {\tt OCNLI}-hypothesis]{   
    \begin{minipage}{5cm}
    \centering    
    \includegraphics[width=.95\textwidth]{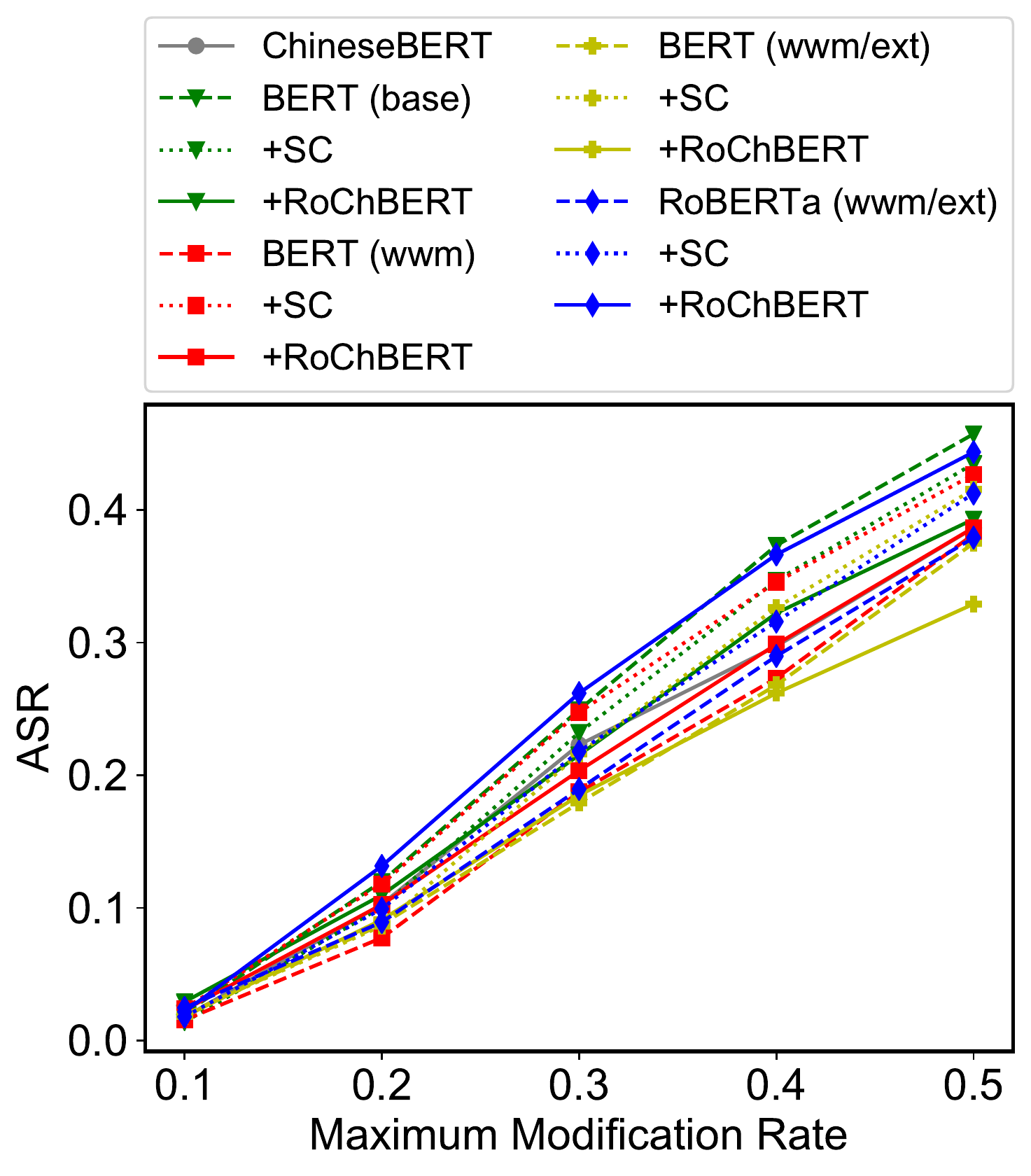}  
    \end{minipage}
    }
    
    \caption{Relation of modification rate and ASR on {\tt OCNLI}.} 
    \label{fig:relation-ocnli} 
    \end{figure*}

\begin{table*}[htbp]
\small
\centering

\begin{tabular}{ccccccc}

\toprule
\textbf{Models}                                                                  & \textbf{Attack} & \textbf{UASR} & \textbf{LASR} & \textbf{MR} & \textbf{Used Samples} & \textbf{Percentage} \\ 
\cmidrule(r){1-1}
\cmidrule(r){2-2}
\cmidrule(r){3-5}
\cmidrule(r){6-7}
\multirow{3}{*}{BERT\textsubscript{base}}  
& PWWS     & 83.62   & 67.66         & 12.96          & 0   & 0\%                 \\
 & TextBuuger & 97.45 & 69.26   & 16.12     & 0   & 0\%         \\
& Random          & 52.77    & 8.19      & 42.85   & 0     & 0\% \\ 
\cmidrule(r){1-1}
\cmidrule(r){2-2}
\cmidrule(r){3-5}
\cmidrule(r){6-7}
\multirow{3}{*}{\begin{tabular}[c]{@{}c@{}}+Adversarial\\ Training\end{tabular}} 
& PWWS     & 70.67     & 50.27     & 19.02   & 9600   & 100\%    \\
& TextBugger   & 96.36   & 64.09  & 16.69  & 9600  & 100\%  \\
& Random     & 36.18   & 16.24    & 30.08     & 9600   & 100\%  \\ 
\cmidrule(r){1-1}
\cmidrule(r){2-2}
\cmidrule(r){3-5}
\cmidrule(r){6-7}
\multirow{3}{*}{+\sysname}   
& PWWS     & 65.18  & 31.63   & 29.49  & 2438  & 25.40\%   \\
& TextBugger   & 64.45   & 34.92  & 20.35  & 1653  & 17.22\%   \\
& Random    & 39.49   & 10.98  & 37.48  & 1588  & 16.54\%    \\ 
\bottomrule
\end{tabular}
\caption{Efficiency of traditional adversarial training and \sysname.}
\label{tab:efficiency}
\end{table*}

\subsection{Impact of Modification Rate}
\label{sec:ap1}
In Figure~\ref{fig:relation-chn},~\ref{fig:relation-dmsc},~\ref{fig:relation-thu} and ~\ref{fig:relation-ocnli} we list all the experiment results about the impact of modification rate with models trained on {\tt ChnSentiCorp}, {\tt DMSC}, {\tt THUCNews} and  {\tt OCNLI} against PWWS, TextBugger and Random attacks.

We can see that as the allowed modification rate increases, ASR is gradually growing. And the ASR of the models with \sysname have the least rise, which proves that even in different settings, the models defended by \sysname still significantly outperform the baselines.

\subsection{Efficiency}
\label{sec:ap_efficiency}
To assess the efficiency of \sysname, we take the BERT\textsubscript{base} fine-tuned on {\tt ChnSentiCorp} dataset as an example and compare \sysname with the traditional adversarial training.
For adversarial training method, we use attack algorithms on the target model with all the training dataset and collect the successful ones as the augmentation dataset.
The results are presented in Table~\ref{tab:efficiency}.
It shows that \sysname only uses 25.40\%, 17.22\% and 16.54\% of the training dataset when conducting PWWS, TextBugger and Random attacks respectively.
Even the adversarial training has used all the training dataset, \sysname still achieves more robust models.
It proves that the data augmentation method used in \sysname is efficient and effective.
It also indicates the key impact of multimodal fusion.

\end{document}